\documentclass[10pt,twocolumn,letterpaper]{article}

\usepackage{cvpr}
\usepackage{times}
\usepackage{epsfig}
\usepackage{graphicx}
\usepackage{amsmath}
\usepackage{amssymb}

\usepackage{graphicx} 
\usepackage{subfigure}

\usepackage{url}            
\usepackage{booktabs}       
\usepackage{amsfonts}       
\usepackage{nicefrac}       
\usepackage{microtype}      
\usepackage{color}
\usepackage{array}

\usepackage{algorithm}
\usepackage{algorithmic}
\usepackage{amssymb}
\usepackage{amsthm}
\usepackage{amsmath,epsfig}
\usepackage{epstopdf}

\usepackage{enumitem}
\usepackage{balance}

\usepackage[pagebackref=true,breaklinks=true,letterpaper=true,colorlinks,bookmarks=false]{hyperref}

 \cvprfinalcopy 

\def\cvprPaperID{3025} 

\begin{document}

\title{ Iterative Normalization: Beyond Standardization towards Efficient Whitening}
\author{Lei Huang \quad Yi Zhou \quad Fan Zhu \quad Li Liu  \quad Ling Shao\\
	Inception Institute of Artificial Intelligence (IIAI), Abu Dhabi, UAE\\
	{\tt\small $\{$lei.huang, yi.zhou,  fan.zhu, li.liu, ling.shao$\}$ @inceptioniai.org}
}
\maketitle
\begin{abstract}
    Batch Normalization (BN) is ubiquitously employed for accelerating neural network training and improving the generalization capability by performing standardization within mini-batches.
    Decorrelated Batch Normalization (DBN) further boosts the above effectiveness by whitening. 
     However, DBN relies heavily on either  a large batch size, or eigen-decomposition that suffers from poor efficiency on GPUs. 
     We propose Iterative Normalization (IterNorm), which employs Newton’s iterations for much more efficient whitening, while simultaneously avoiding the eigen-decomposition. Furthermore, we develop a comprehensive study to show IterNorm has better trade-off between optimization and generalization, with theoretical and experimental support. To this end, we exclusively introduce Stochastic Normalization Disturbance (SND), which measures the inherent stochastic uncertainty of samples when applied to normalization operations.
      With the support of SND, we provide natural explanations to several  phenomena from the perspective of optimization, e.g., why group-wise whitening of DBN generally outperforms full-whitening and why the accuracy of BN degenerates with reduced batch sizes. We demonstrate the consistently improved performance of IterNorm with extensive experiments on CIFAR-10 and ImageNet over BN and DBN. 
\end{abstract}

\section{Introduction}
\label{sec_intro}
\begin{figure}[t]
	\centering
	\centering
	\includegraphics[width=8.4cm]{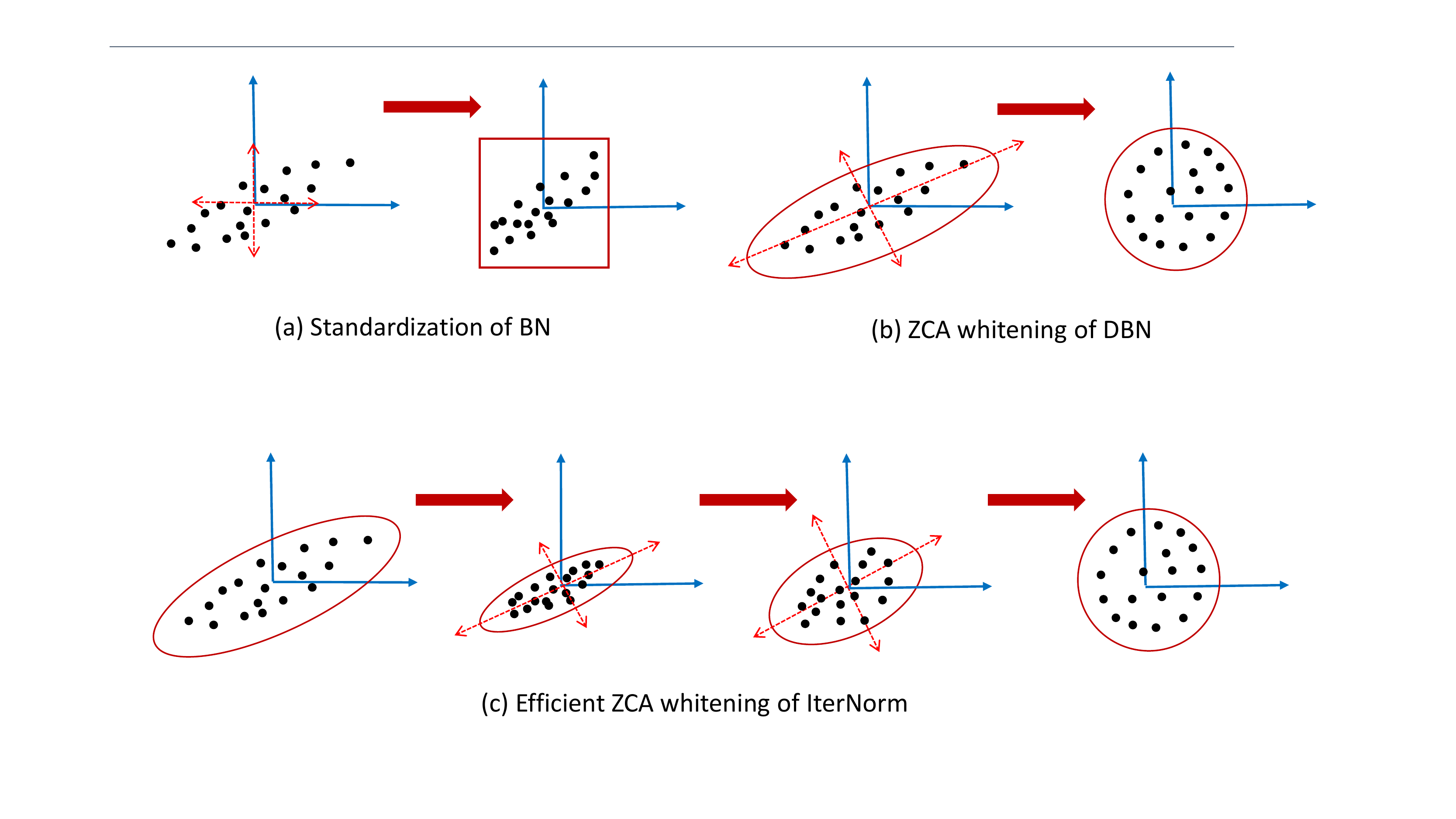}
	\caption{Illustrations of multiple normalization methods on centered data. (a) BN \cite{2015_ICML_Ioffe} performs standardization by stretching/squeezing the data along the axes, such that each dimension has a unit variance; (b) DBN performs ZCA whitening by  stretching/squeezing the data along the eigenvectors, such that the covariance matrix is identical. (c) IterNorm performs efficient ZCA  whitening by progressively adjusting the data along the eigenvectors without eigen-decomposition.  }
	\label{fig:iternorm}
	\vspace{-0.16in}
\end{figure}

Centering, scaling and decorrelating the input data is known as data whitening, which has demonstrated enormous success in speeding up training \cite{1998_NN_Yann}. Batch Normalization (BN) \cite{2015_ICML_Ioffe} extends the operations from the input layer to centering and scaling activations of each intermediate layer within a mini-batch so that each neuron has a zero mean and a unit variance (Figure \ref{fig:iternorm} (a)). BN has been extensively used in various network architectures \cite{2015_CVPR_He,2015_CoRR_Szegedy,
	2016_CoRR_He,2016_CoRR_Zagoruyko,2016_CoRR_Szegedy,2016_CoRR_Huang_a} for its benefits in improving both the optimization efficiency \cite{2015_ICML_Ioffe,2015_NIPS_Desjardins,2018_arxiv_Kohler,2018_NIPS_Bjorck,2018_NIPS_shibani} and generalization capability \cite{2015_ICML_Ioffe,2016_CoRR_Ba,2018_NIPS_Bjorck,2018_ECCV_Wu}. However, instead of performing whitening, BN is only capable of performing standardization, which centers and scales the activations but does not decorrelate them \cite{2015_ICML_Ioffe}. On the other hand, previous works suggest that further decorrelating the activations is beneficial to both the optimization \cite{2015_NIPS_Desjardins,2017_ICML_Luo} and  generalization \cite{2016_ICLR_Cogswell,2016_ICDM_Xiong}. To the end of improving BN with whitening, Decorrelated Batch Normalization (DBN) \cite{2018_CVPR_Huang} is proposed to whiten the activations of each layer within a mini-batch, such that the output of each layer has an isometric diagonal covariance matrix (Figure \ref{fig:iternorm} (b)). DBN improves over BN in regards to both training efficiency and generalization capability, but it relies heavily on a large batch size and eigen-decompositions or singular value decomposition (SVD), which suffers from poor efficiency on GPUs.

In order to address these issues, we propose Iterative Normalization (IterNorm) to further enhance BN with more efficient whitening. IterNorm avoids eigen-decomposition or SVD by employing Newton's iteration for approximating the whitening matrix. Thus, the capacity of GPUs can be effectively exploited. Eigenvalues of the covariance matrix are normalized prior to the iterations with guaranteed convergence condition of Newton's iteration. As illustrated in (Figure \ref{fig:iternorm} (c)), IterNorm stretches the dimensions along the eigenvectors progressively, so that the associated eigenvalues converge to 1 after normalization. One  desirable property is that the convergence speed of IterNorm along the eigenvectors is proportional to the associated eigenvalues \cite{2005_NumerialAlg}. This means the dimensions that correspond to small/zero (when a small batch size is applied) eigenvalues can be largely ignored, given a fixed number of iterations. As a consequence, the sensitivity of IterNorm against batch size can be significantly reduced.



When the data batch is undersized, it is known that the performance of both whitening and standardization on the test data can be significantly degraded \cite{2017_NIPS_Ioffe, 2018_ECCV_Wu}. However, beyond our expectation, we observe that the performance on the training set also significantly degenerates under the same condition. 
We further observe such a phenomenon is caused by the stochasticity introduced by the mini-batch based normalization \cite{2018_ICML_Teye,2018_CVW_Alexander}. 
To allow a more comprehensive understanding and evaluation about the stochasticity, we introduce Stochastic Normalization Disturbance (SND), which is discussed  in Section \ref{sec_analyze_discuss}. With the  support of SND,  we provide a thorough analysis regarding the performance of normalization methods, with respect to the batch size and feature dimensions, and show that IterNorm has better trade-off between optimization and generalization. Experiments on CIFAR-10
~\cite{2009_TR_Alex} and ILSVRC-2012~\cite{2009_ImageNet} demonstrate the consistent improvements of IterNorm over BN and DBN.



\section{Related Work}
\label{sec_relatedWork}

Normalized activations \cite{1998_Schraudolph,2012_AISTATS_Raiko,2012_NN_Gregoire, 2014_ICASSP_Wiesler} have long been known to benefit  neural networks training. 
Some research methodologies attempt to normalize activations by viewing the population statistics as parameters and estimating them directly during training \cite{2012_NN_Gregoire, 2014_ICASSP_Wiesler, 2015_NIPS_Desjardins}. Some of these methods include activations centering in Restricted Boltzmann Machine \cite{2012_NN_Gregoire}/feed-forward neural networks \cite{2014_ICASSP_Wiesler} and activations whitening   \cite{2015_NIPS_Desjardins,2017_ICML_Luo}. This type of normalization may suffer from instability (such as divergence or gradient
explosion) due to 1) inaccurate approximation to the population statistics with local data samples \cite{2014_ICASSP_Wiesler,2015_ICML_Ioffe, 2017_NIPS_Ioffe, 2018_CVPR_Huang} and 2) the internal-covariant shift problem \cite{2015_ICML_Ioffe}.

Ioffe et al., \cite{2015_ICML_Ioffe} propose to perform normalization as a function over mini-batch data and back-propagate through the transformation. 
Multiple standardization options have been discovered for normalizing mini-batch data, including the L2 standardization \cite{2015_ICML_Ioffe}, the L1-standardization \cite{2018_CoRR_Wu,2018_arxiv_Hoffer} and the $L\infty$-standardization  \cite{2018_arxiv_Hoffer}. One critical issue with these methods, however, is that it normally requires a reasonable batch size for estimating the mean and variance. In order to address such an issue, a significant number of standardization approaches are proposed \cite{2016_CoRR_Ba,2018_ECCV_Wu,2017_ICLR_Ren,2018_arxiv_luo,2017_NIPS_Ioffe,2016_axive_Liao,2018_NIPS_Wang,2016_ICASSP_Laurent,2016_CoRR_Cooijmans}. Our work develops in an orthogonal direction to these approaches, and aims at improving BN with decorrelated activations.




Beyond standardization, Huang \etal \cite{2018_CVPR_Huang} propose DBN, which uses ZCA-whitening by eigen-decomposition and back-propagates the transformation. Our approach aims at a much more efficient approximation of the ZCA-whitening matrix in DBN, and suggests that approximating whitening is more effective based on the analysis shown in Section \ref{sec_analyze_discuss}.

Our approach is also related to  works that  normalize the network weights (e.g., either through re-parameterization  \cite{2016_CoRR_Salimans,Huang2017ICCV,2017_Huang_OWN}
or  weight regularization ~\cite{1992_WD_Krogh,2015_NIPS_Neyshabur,2017_ICLR_Pau}), and that specially design either scaling coefficients \& bias values  \cite{2016_ICML_Arpit} or nonlinear function \cite{2017_NIPS_Klambauer}, to normalize activation implicitly \cite{2018_CVW_Alexander}. IterNorm differs from these work in  that it is a data dependent normalization, while  these normalization approaches are independent of the data.

Newton's iteration is also employed in several other deep neural networks. These methods focus on constructing bilinear \cite{2018_BMVC_Lin} or second-order pooling \cite{2018_CVPR_Li} by constraining the power of the covariance matrix and are limited to producing fully-connected activations, while our work provides a generic module that can be ubiquitously built in various neural network frameworks. 
Besides, our method computes the square root inverse of the covariance matrix, instead of  calculating the square root of the covariance matrix \cite{2018_BMVC_Lin, 2018_CVPR_Li}. 


\section{Iterative Normalization}
\label{sec_method}
Let $\mathbf{X} \in \mathbf{R}^{d \times m}$ be a data matrix denoting the mini-batch input  of size $m$ in certain layer.
BN \cite{2015_ICML_Ioffe} works by standardizing the activations over the mini-batch input:

{\setlength\abovedisplayskip{5pt}
	\setlength\belowdisplayskip{5pt}
	\begin{equation}
	\label{eqn:std}
\widehat{\mathbf{X}}=\phi_{Std}(\mathbf{X})
= \Lambda_{std}^{-\frac{1}{2}} (\mathbf{X} - \mathbf{\mu} \cdot \mathbf{1}^T),
	\end{equation}
}
\hspace{-0.08in}where $\mathbf{\mu} = \frac{1}{m} \mathbf{X} \cdot \mathbf{1}$ is the
mean of $\mathbf{X}$,  $\Lambda_{std}=\mbox{diag}(\sigma_1^2, \ldots,\sigma_d^2) + \epsilon \mathbf{I}$, $\sigma_i^2$ is the dimension-wise variance corresponding to the \emph{i}-th dimension, $\mathbf{1}$ is a column vector of all ones, and $\epsilon>0$ is a small number to prevent numerical instability.
Intuitively, standardization ensures that the normalized output gives equal importance to each dimension by multiplying the scaling matrix $\Lambda_{std}^{-\frac{1}{2}}$ (Figure \ref{fig:iternorm} (a)).

DBN \cite{2018_CVPR_Huang} further uses ZCA whitening  to produce the whitened output as\footnote{DBN and BN both use learnable dimension-wise scale and shift parameters to recover the possible loss of representation capability.}:
 \begin{small}
\begin{equation}
\label{eqn:whiten}
\phi_{ZCA}(\mathbf{X})
= \mathbf{D}\Lambda^{-\frac{1}{2}} \mathbf{D}^T (\mathbf{X} - \mathbf{\mu} \cdot \mathbf{1}^T),
\end{equation}
\end{small}
\hspace{-0.05in}where  $\Lambda=\mbox{diag}(\sigma_1, \ldots,\sigma_d)$ and $\mathbf{D}=[\mathbf{d}_1, ...,
\mathbf{d}_d]$ are the eigenvalues and associated eigenvectors of $\Sigma$, \ie $\Sigma = \mathbf{D}
\Lambda \mathbf{D}^T$.   $\Sigma = \frac{1}{m} (\mathbf{X} - \mathbf{\mu}
\cdot \mathbf{1}^T) (\mathbf{X} - \mathbf{\mu} \cdot \mathbf{1}^T)^T +
\epsilon \mathbf{I}$ is the covariance matrix of the centered input.
ZCA whitening works by stretching or squeezing  the dimensions along the eigenvectors
such that the associated eigenvalues to be $1$ (Figure \ref{fig:iternorm} (b)). Whitening the activation ensures that all dimensions along the eigenvectors have equal importance in the subsequent linear layer.

One crucial problem of ZCA whitening is that calculating the whitening matrix requires eigen-decomposition or  SVD, as shown in Eqn. \ref{eqn:whiten}, which heavily constrains its practical applications.
We observe that Eqn. \ref{eqn:whiten}  can be viewed as the square root inverse of the covariance matrix denoted by $\Sigma^{-\frac{1}{2}}$, which multiplies the centered input. The square root inverse of one specific matrix can  be calculated using Newton's iteration methods \cite{2005_NumerialAlg}, which avoids executing eigen-decomposition or SVD.

\begin{algorithm}[tb]
	\caption{Whitening activations with Newton's iteration.}
	\label{alg_forward}
	\begin{algorithmic}[1]
		\begin{small}
			\STATE \textbf{Input}: mini-batch inputs $ \mathbf{X} \in \mathbb{R}^{d \times m} $.
			\STATE \textbf{Hyperparameters}: $\epsilon$, iteration number $T$.
			\STATE \textbf{Output}: the ZCA-whitened activations $ \widehat{\mathbf{X}} $.
			\STATE	calculate mini-batch mean: $\mathbf{\mu} = \frac{1}{m} \mathbf{X} \cdot \mathbf{1}$.
			\STATE	calculate centered activation: $\mathbf{X}_C = \mathbf{X}-\mathbf{\mu}  \cdot \mathbf{1}^T $.
			\STATE	calculate covariance matrix: $\Sigma = \frac{1}{m}\mathbf{X}_C \mathbf{X}_C ^T + \epsilon \mathbf{I}$.
			\STATE	calculate trace-normalized covariance matrix $\Sigma_{N}$ by Eqn .\ref{eqn:Spectral_Norm}.
			\STATE $\mathbf{P}_0=\mathbf{I}$.
			\FOR {$k = 1 ~~to ~~T $}
			\STATE $\mathbf{P}_{k}=\frac{1}{2} (3 \mathbf{P}_{k-1} - \mathbf{P}_{k-1}^{3} \Sigma_{N})$
			\ENDFOR	
			\STATE  calculate whitening matrix: $\Sigma^{-\frac{1}{2}} = \mathbf{P}_T / \sqrt{tr(\Sigma)}$.
			\STATE  calculate  whitened output: $\widehat{\mathbf{X}} = \Sigma^{-\frac{1}{2}} \mathbf{X}_C $.
%
		\end{small}
	\end{algorithmic}
\end{algorithm}

\subsection{Computing $\Sigma^{-\frac{1}{2}}$ by Newton's Iteration}
 Given the square matrix $\mathbf{A}$, Newton's method calculates $\mathbf{A}^{-\frac{1}{2}}$ by the following iterations \cite{2005_NumerialAlg}:
  \begin{small}
\begin{equation}
\label{eqn:Iteration}
\begin{cases}
\mathbf{P}_0=\mathbf{I} \\
\mathbf{P}_{k}=\frac{1}{2} (3 \mathbf{P}_{k-1} - \mathbf{P}_{k-1}^{3} \mathbf{A}), ~~ k=1,2,...,T,
\end{cases}
\end{equation}
\end{small}
\hspace{-0.05in}where $T$ is the iteration number. $\mathbf{P}_{k}$ will be converged to  $\mathbf{A}^{-\frac{1}{2}}$ under the condition $\| \mathbf{A} -\mathbf{I} \|_2 < 1$.

In terms of applying Newton's methods to calculate the inverse square root  of the covariance matrix  $\Sigma^{-\frac{1}{2}}$, one crucial problem  is  $\Sigma$ cannot be guaranteed to satisfy the convergence condition  $\| \Sigma -\mathbf{I} \|_2 < 1$. That is because $\Sigma$ is calculated over mini-batch samples and thus varies during training. If the convergence condition cannot be perfectly satisfied, the training can be highly instable \cite{2005_NumerialAlg,2018_CVPR_Li}. 
To address this issue, we observe that one sufficient condition for convergence is to ensure the eigenvalues of the covariance matrix are less than $1$.  We thus propose to construct a transformation $\Sigma_{N}=F(\Sigma)$ such that $\| \Sigma_{N} \|_2 < 1$, and ensure the transformation is differentiable such that the gradients can back-propagate through this transformation. One feasible transformation is to  normalize the eigenvalue as follows:
  \begin{small}
\begin{eqnarray}
\label{eqn:Spectral_Norm}
\Sigma_{N} = \Sigma / tr(\Sigma),
\end{eqnarray}
  \end{small}
\hspace{-0.05in}where $tr(\Sigma)$ indicates the trace of $\Sigma$. Note that $\Sigma_{N}$ is also a semi-definite matrix and thus all of its eigenvalues are greater than or equal to $0$. Besides, $\Sigma_{N}$ has the property that the sum of its eigenvalues is  $1$. Therefore,  $\Sigma_{N}$ can surely satisfy the convergence condition.  We can thus calculate  the inverse  square root  $\Sigma_{N}^{-\frac{1}{2}}$ by Newton's method as Eqn. \ref{eqn:Iteration}. Given $\Sigma_{N}^{-\frac{1}{2}}$ , we can compute $\Sigma^{-\frac{1}{2}}$ based on Eqn. \ref{eqn:Spectral_Norm}, as follows:
\begin{small}
\begin{eqnarray}
\label{eqn:inv_Spectral_Norm}
\Sigma^{-\frac{1}{2}} = \Sigma_{N}^{-\frac{1}{2}} / \sqrt{tr(\Sigma)}.
\end{eqnarray}
\end{small}
\hspace{-0.05in}Given $\Sigma^{-\frac{1}{2}}$, it's easy to  whiten the activations by multiplying $\Sigma^{-\frac{1}{2}}$ with the centered inputs.  In summary, Algorithm \ref{alg_forward} describes our proposed methods for whitening the activations in neural networks.

Our method first normalizes the eigenvalues of the covariance matrix, such that the convergence condition of Newton's iteration is satisfied. We then progressively stretch the dimensions along the eigenvectors, such that the final associate eigenvalues are all ``$1$'', as shown in Figure \ref{fig:iternorm} (c). Note that the speed of convergence of the eigenvectors is proportional to the associated eigenvalues \cite{2005_NumerialAlg}. That is, the larger the eigenvalue is, the faster its associated dimension along the eigenvectors converges.  Such a mechanism is a remarkable property to control the extent of  whitening, which is essential for the success of whitening activations, as pointed out in \cite{2018_CVPR_Huang}, and will be further discussed in Section \ref{sec_analyze_discuss}.

\subsection{Back-propagation}
As pointed out by \cite{2015_ICML_Ioffe, 2018_CVPR_Huang}, viewing standardization or whitening as functions over the mini-batch data and back-propagating through the normalized transformation are essential for stabilizing training. Here, we derive the back-propagation pass of IterNorm. Denoting $L$ as the loss function, 
the key is to calculate $\frac{\partial{L}}{\partial{\Sigma}}$, given $\frac{\partial{L}}{\partial{\Sigma^{-1/2}}}$. Let's denote $\mathbf{P}_{T} = \Sigma_N^{-\frac{1}{2}}$, where $T$ is the iteration number.
Based on the chain rules, we have:
\begin{small}
\begin{eqnarray}
\label{eqn:backward-1}
\frac{\partial{L}}{\partial{\mathbf{P}_{T}}}&=& \frac{1}{\sqrt{tr(\Sigma)}} \frac{\partial{L}}{\partial{\Sigma^{-\frac{1}{2}}}} \nonumber \\
\frac{\partial{L}}{\partial{\Sigma_{N}}}&=& -\frac{1}{2} \sum_{k=1}^{T} (\mathbf{P}_{k-1}^3)^T \frac{\partial{L}}{\partial{\mathbf{P}_k}}  \nonumber  \\
\frac{\partial{L}}{\partial{\Sigma}} &=&  \frac{1}{tr(\Sigma)} \frac{\partial{L}}{\partial{\Sigma_{N}}}
-\frac{1}{(tr(\Sigma))^2} tr(\frac{\partial{L}}{\partial{ \Sigma_{N}}}^T \Sigma)   \mathbf{I}   \nonumber \\
~~~&-& \frac{1}{2(tr(\Sigma))^{3/2}} tr((\frac{\partial{L}}{\partial{\Sigma^{-1/2}}})^T \mathbf{P}_T) \mathbf{I},
\end{eqnarray}
\end{small}
\hspace{-0.05in}where $\frac{\partial{L}}{\partial{\mathbf{P}_{k}}}$ can be calculated by following iterations:
\begin{small}

\begin{align}
\label{eqn:backward-iteration}
\frac{\partial{L}}{\partial{\mathbf{P}_{k-1}}}& =\frac{3}{2} \frac{\partial{L}}{\partial{\mathbf{P}_{k}}}
-\frac{1}{2} \frac{\partial{L}}{\partial{\mathbf{P}_{k}}}  (\mathbf{P}_{k-1}^2 \Sigma_{N})^T
-\frac{1}{2}  (\mathbf{P}_{k-1}^2)^T  \frac{\partial{L}}{\partial{\mathbf{P}_{k}}} \Sigma_{N}^T
\nonumber \\
&-  \frac{1}{2}(\mathbf{P}_{k-1})^T \frac{\partial{L}}{\partial{\mathbf{P}_{k}}} (\mathbf{P}_{k-1} \Sigma_{N})^T
,  ~~k=T,...,1.
\end{align}
\end{small}

Note that in Eqn. \ref{eqn:Spectral_Norm} and \ref{eqn:inv_Spectral_Norm},  $tr(\Sigma)$ is a function for mini-batch examples and is needed to back-propagate through it to stabilize the training.
Algorithm \ref{alg_backprop} summarizes the back-propagation pass of our proposed IterNorm.
More details of back-propagation derivations are shown in Appendix \ref{sec:appendix_back}.

\begin{algorithm}[tb]
	\caption{The respective backward pass of Algorithm \ref{alg_forward}.}
	\label{alg_backprop}
	\begin{algorithmic}[1]
		\begin{small}
			\STATE \textbf{Input}: mini-batch  gradients respect to whitened activations: $ \frac{\partial L}{\partial \widehat{\mathbf{X}}} $.
			auxiliary data from respective forward pass: (1) $\mathbf{X}_C$; (2) $\Sigma^{-\frac{1}{2}}$; (3) $\{\mathbf{P}_k\}$.
			
			\STATE \textbf{Output}: the gradients with respect to the inputs: $ \frac{\partial L}{\partial \mathbf{X}} $.
			\STATE calculate the gradients with respect to $\Sigma^{-\frac{1}{2}}$: $\frac{\partial L}{\partial \Sigma^{-\frac{1}{2}}}	 =\frac{\partial L}{\partial \widehat{\mathbf{X}}}     \mathbf{X}_C^T$.
			\STATE calculate $\frac{\partial L}{\partial \Sigma}$ based on Eqn. \ref{eqn:backward-1} and \ref{eqn:backward-iteration}.
			\STATE calculate: $\mathbf{f}=\frac{1}{m} \frac{\partial L}{\partial \widehat{\mathbf{X}}} \cdot \mathbf{1} $.
			\STATE calculate: $\frac{\partial L}{\partial \mathbf{X}} =  \Sigma^{-\frac{1}{2}} (\frac{\partial L}{\partial \widehat{\mathbf{X}}}-\mathbf{f} \cdot \mathbf{1}^T ) +  \frac{1}{m}  (\frac{\partial L}{\partial \Sigma} + \frac{\partial L}{\partial \Sigma}^T) \mathbf{X}_C $.
		\end{small}
	\end{algorithmic}
\end{algorithm}

\subsection{Training and Inference}
\label{sec:trainingAndInfer}
Like the previous normalizing activation methods \cite{2015_ICML_Ioffe,2016_CoRR_Ba,2018_CVPR_Huang,2018_ECCV_Wu}, our IterNorm can be used as a module and inserted into a network extensively.
Since IterNorm is also a method for mini-batch data, we  use the running average to calculate the population mean $\hat{\mu}$ and whitening matrix $\widehat{\Sigma}^{-\frac{1}{2}}$, which is used during inference.
Specifically, during training, we initialize $\hat{\mu}$ as
$\mathbf{0}$ and $\widehat{\Sigma}^{-\frac{1}{2}}$ as $\mathbf{I}$ and update them as follows:
 \begin{small}
\begin{eqnarray}
\label{eqn:running average}
\hat{\mu} &=& (1-\lambda) ~\hat{\mu} + \lambda ~\mathbf{\mu} \nonumber \\
\widehat{\Sigma}^{-\frac{1}{2}} & = &(1-\lambda) \widehat{\Sigma}^{-\frac{1}{2}} + \lambda \Sigma^{-\frac{1}{2}},
\end{eqnarray}
\end{small}
\hspace{-0.05in}where $\mathbf{\mu}$ and $\Sigma^{-\frac{1}{2}}$ are the mean and whitening matrix calculated within each mini-batch during training, and $\lambda$ is the momentum of  running average.

Additionally, we also use the extra learnable parameters $\gamma$
and $\beta$, as in previous normalization methods \cite{2015_ICML_Ioffe,2016_CoRR_Ba,2018_CVPR_Huang,2018_ECCV_Wu}, since
normalizing the activations constrains the model's capacity for
representation.  Such a process has been shown to be effective \cite{2015_ICML_Ioffe,2016_CoRR_Ba,2018_CVPR_Huang,2018_ECCV_Wu}.
\vspace{-0.12in}
\paragraph{Convolutional Layer} For a CNN, the input is $\mathbf{X}_C \in \mathbb{R}^{h \times w \times d
	\times m} $, where $h$ and $w$ indicate the height and width of the
feature maps, and $d$ and $m$ are the number of feature maps and examples, respectively.
Following \cite{2015_ICML_Ioffe}, we view each spatial position of the
feature map as a sample. We thus unroll $\mathbf{X}_C$ as $\mathbf{X}
\in \mathbb{R}^{ d \times (m h w)}$ with $ m \times h \times w$ examples and $d$
feature maps. The whitening operation is performed over the unrolled
$\mathbf{X}$.

	\vspace{-0.12in}
 \paragraph{Computational Cost} The main computation of our IterNorm includes  calculating the covariance matrix, the iteration operation and the whitened output. The computational costs of the first and the third operation are equivalent to the $1\times 1$ convolution. The second operation's computational cost is $T d^3$.  Our method is comparable to the convolution operation. To be specific, given the internal activation $\mathbf{X}_C \in \mathbb{R}^{h \times w \times d  	\times m} $, the $3 \times 3$ convolution with the same input and output feature maps costs $9hwmd^2$, while our IterNorm costs $2hwmd^2 + T d^3$. The relative cost of IterNorm for $3 \times 3$ convolution is $2/9 + Td/mhw$. 
 Further, we can use group-wise whitening, as introduced in \cite{2018_CVPR_Huang} to improve the efficiency when the dimension $d$ is large. We also compare the wall-clock time of IterNorm, DBN  \cite{2018_CVPR_Huang}  and $3 \times 3$ convolution in Appendix \ref{sec:appendix_time}.

 During inference, IterNorm can be viewed as a $1\times 1$ convolution and merged to  adjacent  convolutions. Therefore, IterNorm does not introduce any extra costs in memory or computation during inference.

\section{Stochasticity of Normalization}
\label{sec_analyze_discuss}


Mini-batch based normalization methods are sensitive to the batch size \cite{2017_NIPS_Ioffe,2018_CVPR_Huang,2018_ECCV_Wu}.
 As described in \cite{2018_CVPR_Huang}, fully whitening the activation may suffer from degenerate performance while the number of data in a mini-batch is not sufficient.  
They \cite{2018_CVPR_Huang} thus propose to use group-wise whitening  \cite{2018_CVPR_Huang}.
 Furthermore, standardization also suffers from degenerated performance under the scenario of micro-batch \cite{2018_NIPS_Wang}. 
 These works argue that undersized data batch makes the estimated population statistics highly noisy, which results in a degenerating performance during  inference  \cite{2016_CoRR_Ba,2017_NIPS_Ioffe,2018_ECCV_Wu}.


In this section, we will provide a more thorough analysis regarding  the performance of normalization methods, with respect to the batch size and feature dimensions. We show that normalization (standardization or whitening) with undersized data batch not only suffers from degenerate performance during inference, but also  encounter the difficulty in optimization during training. This is caused by the Stochastic Normalization Disturbance (SND), which we will describe.

\subsection {Stochastic Normalization Disturbance}
\begin{figure}[]
	\centering
	\subfigure[batch size of 16]{
		\begin{minipage}[c]{.46\linewidth}
			\centering
			\includegraphics[width=4.0cm]{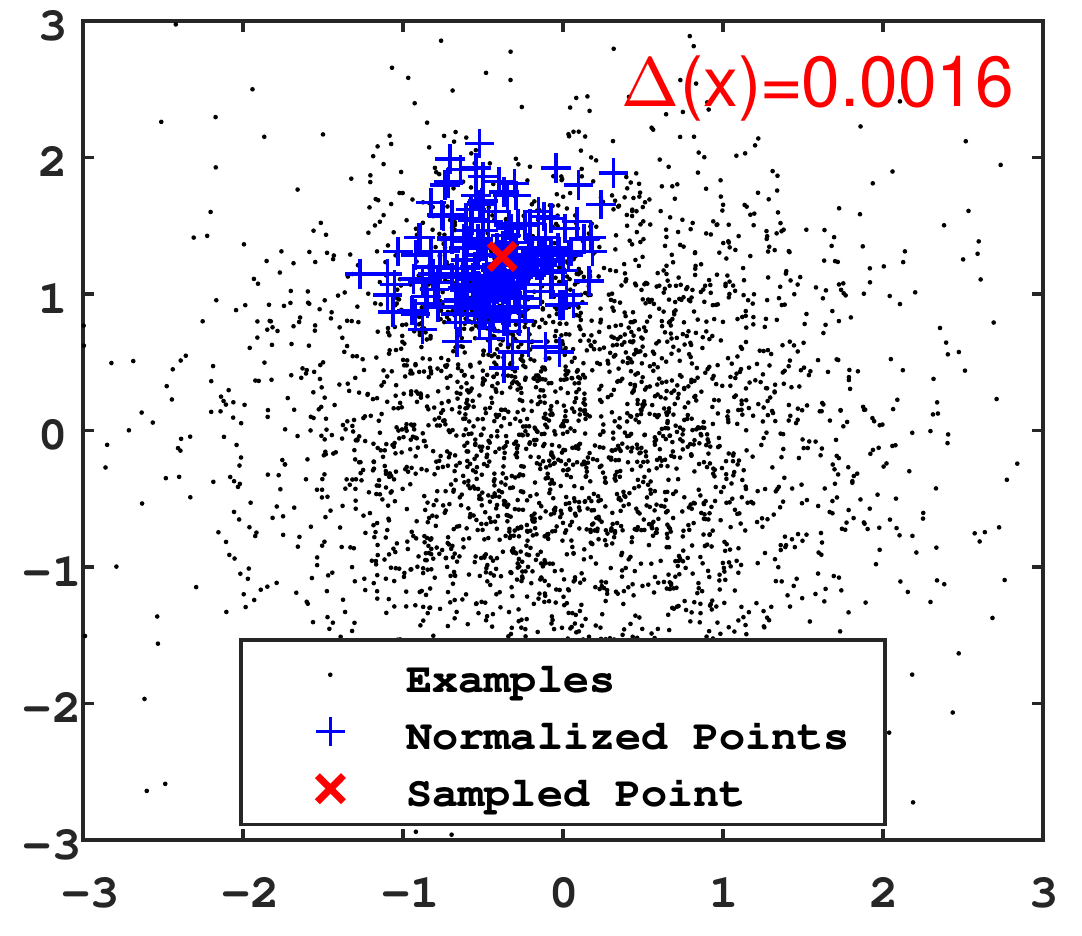}
		\end{minipage}
	}
	\subfigure[batch size of 64]{
		\begin{minipage}[c]{.46\linewidth}
			\centering
			\includegraphics[width=4.0cm]{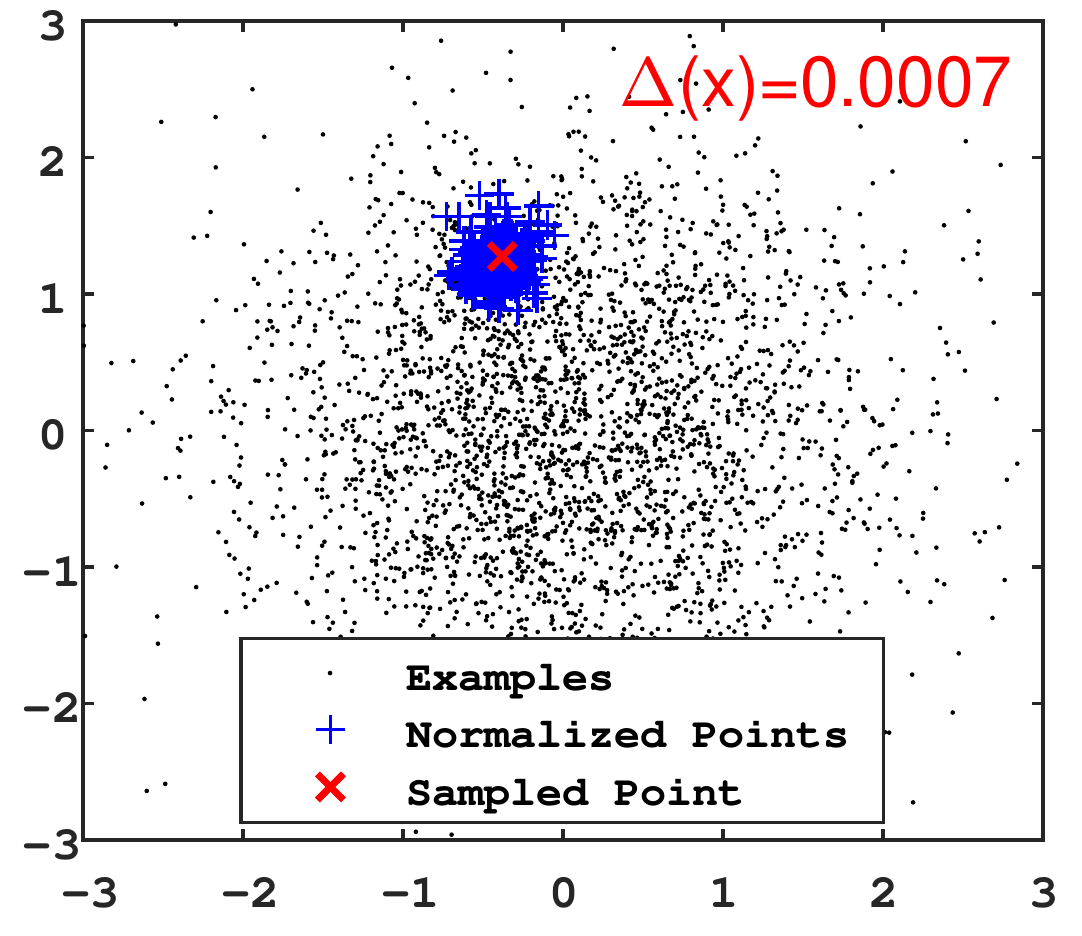}
		\end{minipage}
	}
	\caption{Illustration of SND with different batch sizes. We sample 3000 examples (black points) from Gaussian distribution. We show a given example $\mathbf{x}$ (red cross) and its BN outputs  (blue plus sign), when normalized over different sample sets $\mathbf{X}^B$. (a) and (b) show the results with batch sizes $B$ of 16 and 64, respectively.}
	\label{fig:stochasticy}
	\vspace{-0.16in}
\end{figure}

Given a sample $\mathbf{x} \in \mathbb{R}^d$ from a distribution $P_{\chi}$, we take a sample set $\mathbf{X}^B=\{\mathbf{x}_1,...,\mathbf{x}_B, \mathbf{x}_i \sim P_{\chi} \}$ with a size of $B$. We denote the normalization operation as $F(\cdot)$  and the normalized output as $\hat{\mathbf{x}}=F( \mathbf{X}^B ;\mathbf{x})$.
For a certain $\mathbf{x}$,  $\mathbf{X}^B$ can be viewed as a random variable \cite{2018_CoRR_Andrei,2018_ICML_Teye}. $\hat{\mathbf{x}}$ is thus a random variable which shows the stochasticity.  It's interesting to explore the statistical  momentum of $\mathbf{x}$ to measure the magnitude of the stochasticity. Here we define the \emph{Stochastic Normalization Disturbance} (SND) for the sample $\mathbf{x}$ over the normalization $F(\cdot)$ as:
 \begin{small}
 	\begin{eqnarray}
 	\label{eqn:SND}
 \mathbf{\Delta}_{F}(\mathbf{x})=\mathbf{E}_{\mathbf{X^B}} (\| \hat{\mathbf{x}}- \mathbf{E}_{\mathbf{X}^B} (\hat{\mathbf{x}})  \|_2).
 	\end{eqnarray}
 \end{small}
\hspace{-0.05in}It's difficult to accurately compute this momentum if no further assumptions are made over the random variable $\mathbf{X}^B$, however, we can explore its empirical estimation over the sampled sets as follows:
 \begin{small}
 	\begin{eqnarray}
 	\label{eqn:empirical_SND}
       \widehat{\mathbf{\Delta}}_{F}(\mathbf{x})=\frac{1}{s} \sum_{i=1}^{s} \| F( \mathbf{X}^B_i ; \mathbf{x}) - \frac{1}{s} \sum_{j=1}^s F( \mathbf{X}^B_j ; \mathbf{x})     \|,
 	\end{eqnarray}
 \end{small}
\hspace{-0.05in}where $s$ denotes the time of sampling.
 Figure \ref{fig:stochasticy} gives the illustration of sample $\mathbf{x}$'s SND with respect to the operation of BN.
 We can find that SND is closely related to the batch size. When batch size is large, the given sample $\mathbf{x}$ has a small value of SND and the transformed outputs have a compact distribution. As a consequence,   the stochastic uncertainty $\mathbf{x}$ can be low.

 SND can be used  to evaluate the stochasticity of  a sample after the normalization operation, which works like the dropout rate \cite{2014_JMLR_Nitish}.
 We can further define the normalization operation $F(\cdot)$'s SND as: $\mathbf{\Delta}_{F}= \mathbf{E}_{\mathbf{x}} ( \mathbf{\Delta}(\mathbf{x}))$ and it's empirical estimation  as $\widehat{\mathbf{\Delta}}_{F}=\frac{1}{N}\sum_{i=1} ^{N} \widehat{\mathbf{\Delta}}(\mathbf{x})$ where $N$ is the number of sampled examples.  $\mathbf{\Delta}_{F}$ describes the magnitudes  of stochasticity for corresponding normalization operations.

Exploring the exact statistic behavior of SND  is difficult and out of the scope of this paper. We can, however, explore the relationship of SND  related to the batch size and feature dimension. We find that our defined SND gives a reasonable explanation to why we should control the extent of whitening and
 why mini-batch based normalizations have a degenerate performance when given a small batch size.

\subsection{Controlling the Extent of Whitening}
\begin{figure}[t]
	\centering
\hspace{-0.2in}	\subfigure[batch size of 60,000]{
		\begin{minipage}[c]{.44\linewidth}
			\centering
			\includegraphics[width=4.1cm]{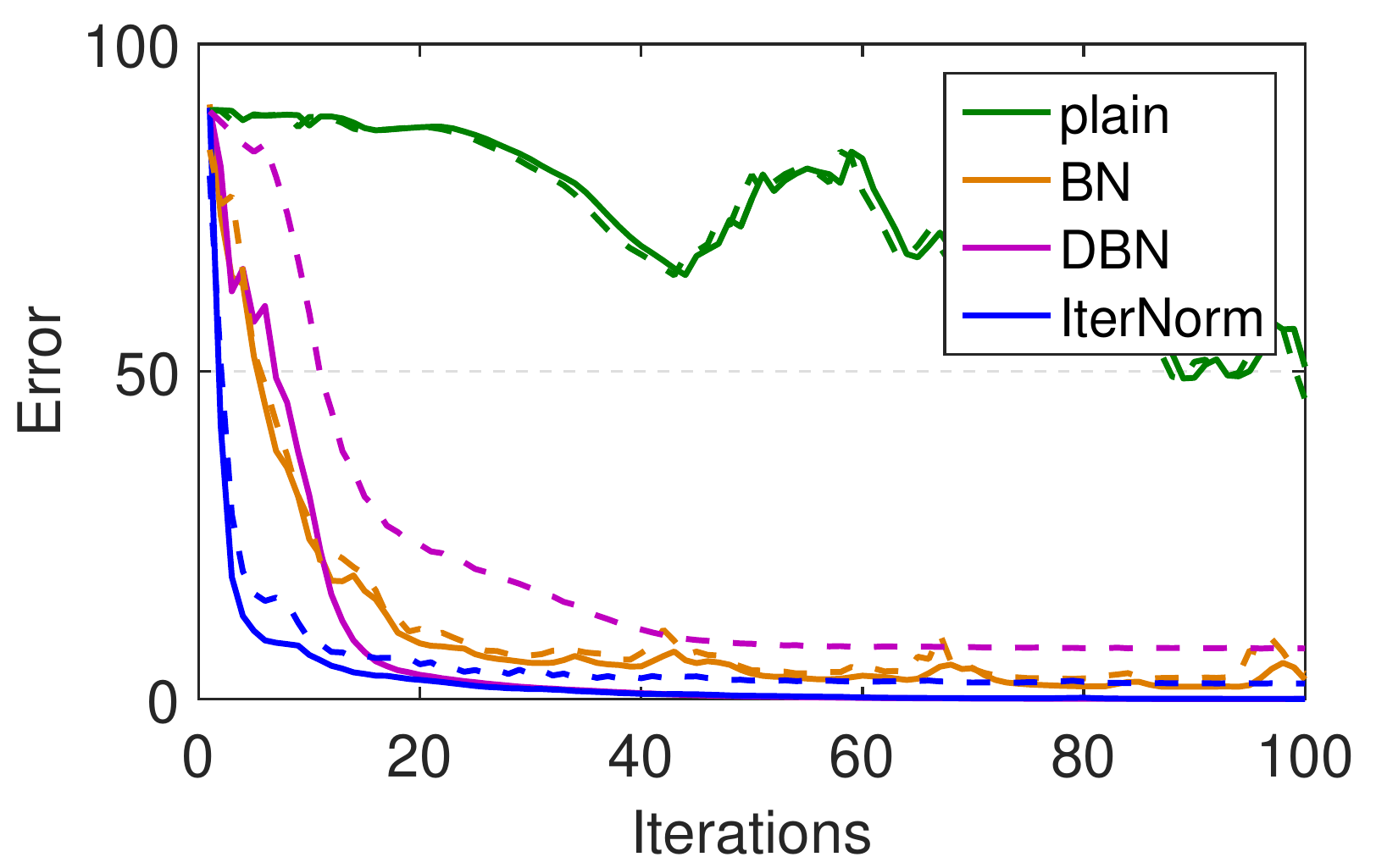}
		\end{minipage}
	}
\hspace{0.1in}	\subfigure[batch size of 2]{
		\begin{minipage}[c]{.44\linewidth}
			\centering
			\includegraphics[width=4.1cm]{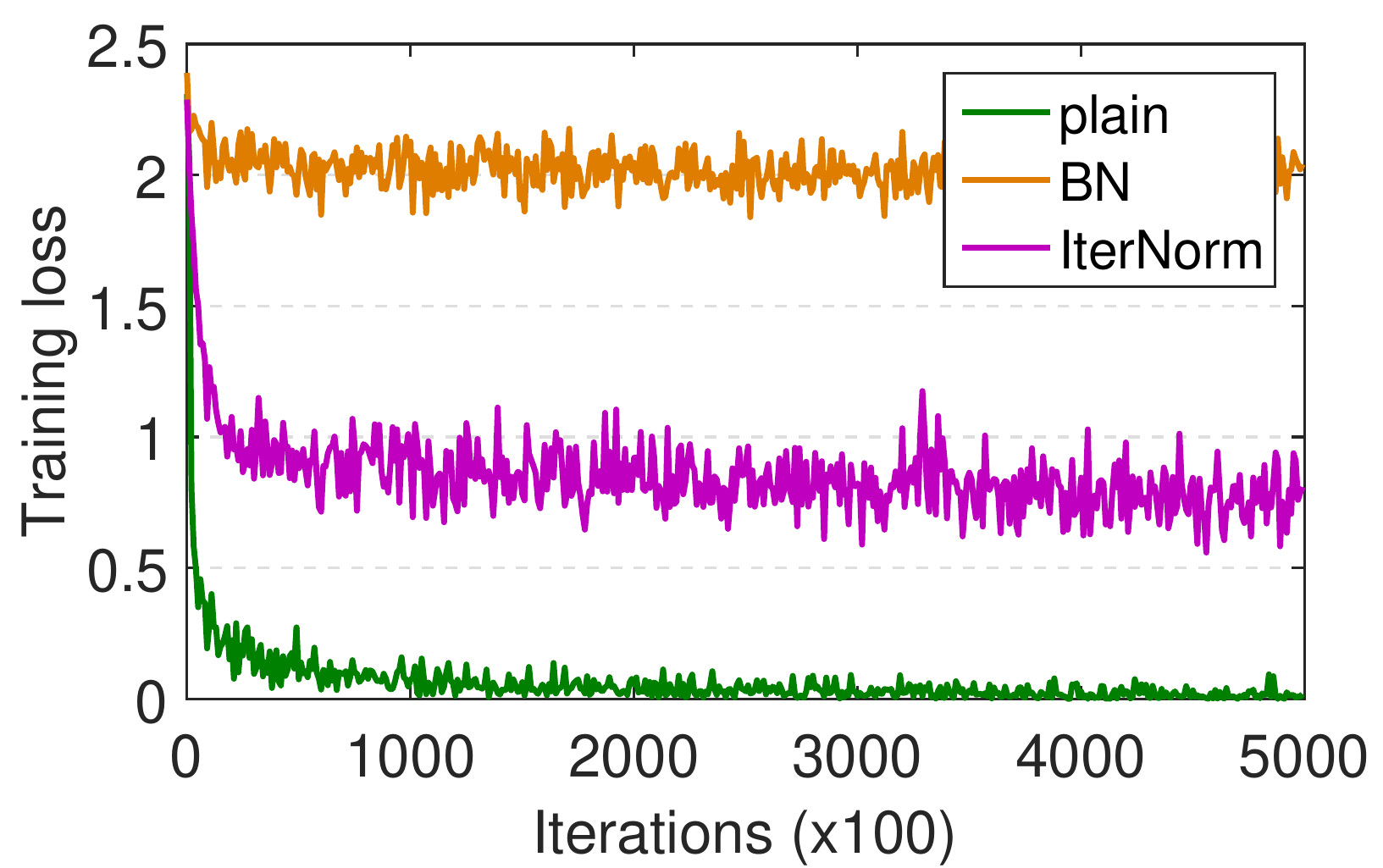}
		\end{minipage}
	}
	\caption{Ablation study in training a 4-layer MLP on MNIST. The number of neurons in each hidden layer is 100. We use full batch gradient and report the best results with respect to the training loss among learning rates=$\{0.2,0.5,1,2,5\}$, and stochastic gradient with batch size of 2 among learning rates= $\{0.005,0.01,0.02, 0.05,0.1\}$. (a) shows the training (solid lines) and test (dashed lines) errors with respect to the iterations, and (b) shows the training loss. `plain' is referred to as the network without normalization. }
	\label{fig:MNIST_Experiment}
	\vspace{-0.16in}
\end{figure}

We start with experiments on multi-layer perceptron (MLP) over MNIST dataset, by using the full batch gradient (batch size =60,000), as shown in Figure \ref{fig:MNIST_Experiment} (a).  We find that all normalization methods significantly improve the performance. One interesting observation is that full-whitening the activations with such a large batch size still underperforms the approximate-whitening of IterNorm, in terms of training efficiency.
Intuitively, full-whitening the activations may lead to amplifying the dimension with small eigenvalues, which may correspond to the noise. Exaggerating this noise may be harmful to  learning, especially lowering down the generalization capability as shown in Figure \ref{fig:MNIST_Experiment} (a) that DBN has diminished test performance.  We provide further analysis based on SND, along with the conditioning analysis. 
It has been shown that improved conditioning can accelerate training \cite{1998_NN_Yann,2015_NIPS_Desjardins}, while increased stochasticity can slow down training  but likely to improve generalization \cite{2014_JMLR_Nitish} .

\begin{figure}[]
 	\centering
 \hspace{-0.2in}	\subfigure[Condition number]{
 		\begin{minipage}[c]{.44\linewidth}
 			\centering
 			\includegraphics[width=4.2cm]{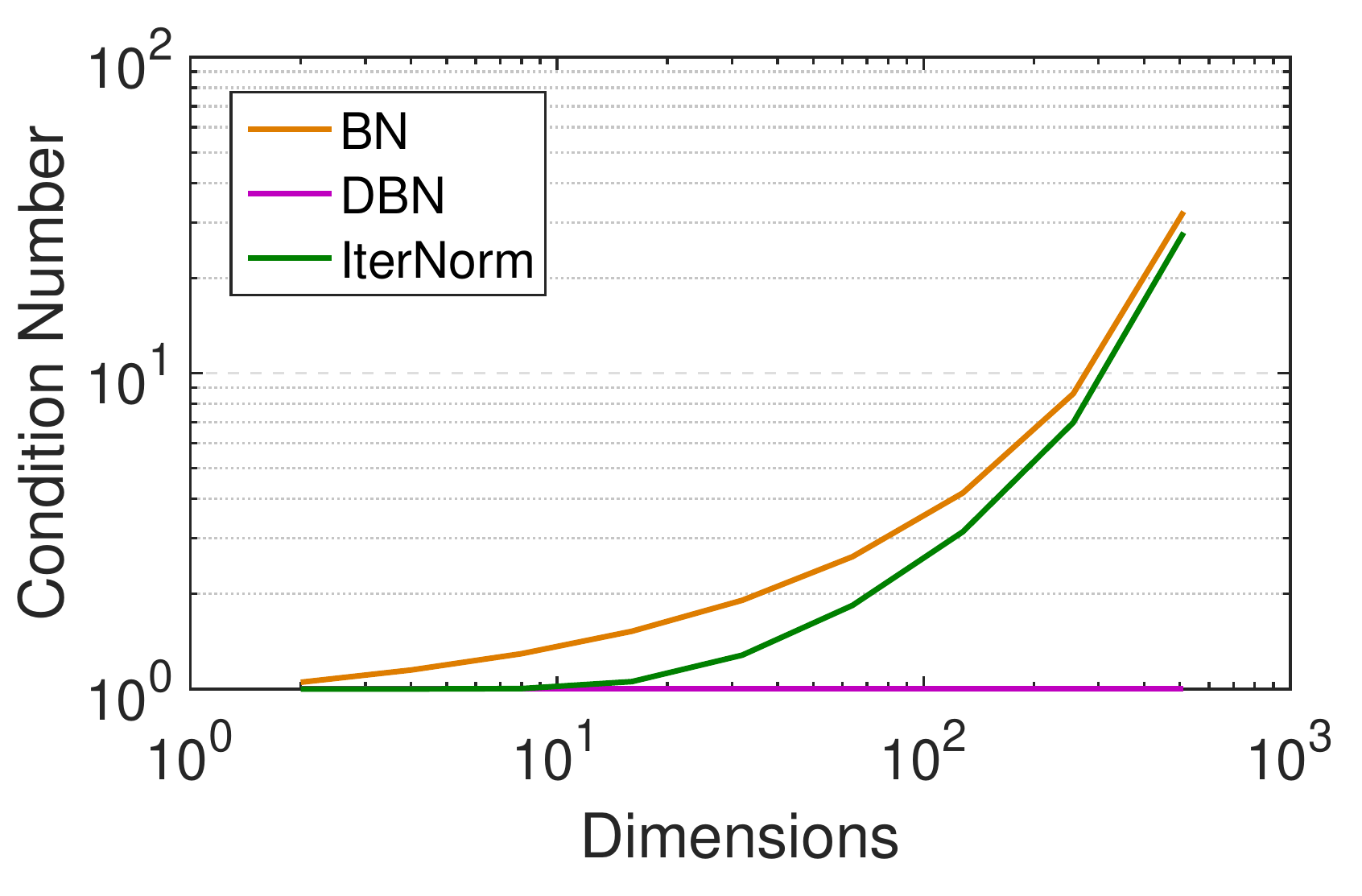}
 		\end{minipage}
 	}
\hspace{0.05in} 	\subfigure[SND]{
 		\begin{minipage}[c]{.44\linewidth}
 			\centering
 			\includegraphics[width=4.2cm]{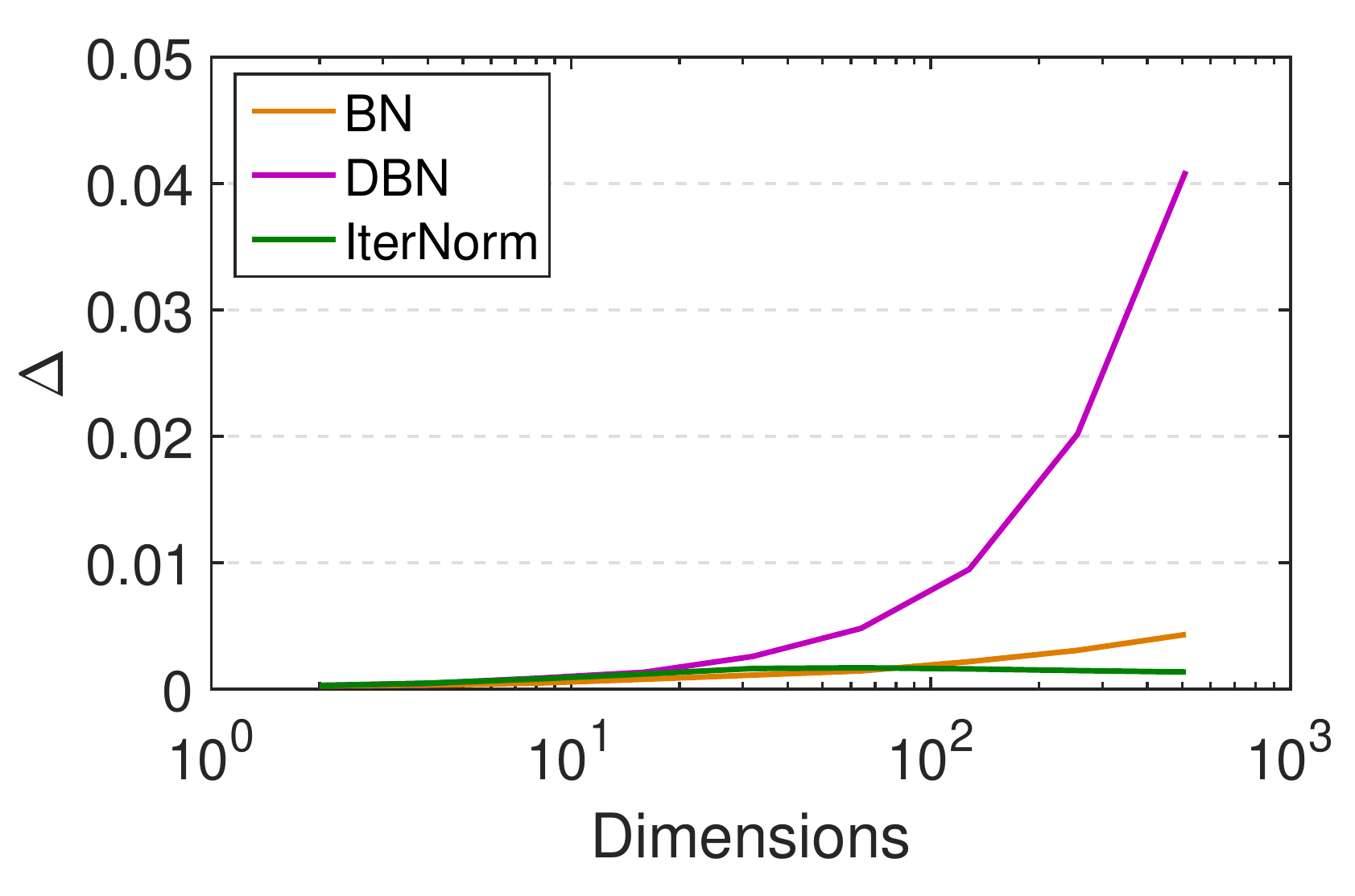}
 		\end{minipage}
 	}
 	\caption{Comparison of different normalization operations in condition number of covariance matrix of normalized output (a) and SND (b). We sample 60,000 examples from Gaussian distribution and choose a batch size of $1024$, and observe the results with respect to the dimensions from $2^1$ to $2^9$, averaged over 10 times. }
 	\label{fig:LargeBatch}
 	\vspace{-0.2in}
 \end{figure}

We experimentally explore the consequent effects of improved conditioning \cite{1998_NN_Yann} with SND through BN (standardization), DBN (full-whitening) and IterNorm (approximate-whitening with 5 iterations).  We calculate the condition number of covariance matrix of normalized output, and the SND for different normalization methods (as shown in Figure \ref{fig:LargeBatch}).
We find that DBN has the best conditioning with an exact condition number as 1, however it significantly enlarges SND, especially in a high-dimensional space. 
Therefore, full-whitening can not consistently improve the training efficiency, even for the highly improved conditioning, which is balanced out by the larger SND. 
Such an observation also explains why group-based whitening \cite{2018_CVPR_Huang} (by reducing the number of dimensions that will be whitened) works better from the training perspective.  

IterNorm has consistently improved conditioning over different dimensions compared to BN.  Interestingly, IterNorm has a reduced SND in a high-dimensional space, since it can adaptively normalize the dimensions along different eigenvalues based on the convergence theory of Newton's iteration \cite{2005_NumerialAlg}. Therefore, IterNorm possesses a better trade-off between the improved conditioning and SND, which naturally illustrates IterNorm can be more efficiently trained.
We also provide the results of IterNorm when applying different iteration numbers in Appendix \ref{sec:appendix_iteration}.

\subsection{Micro-batch Problem of BN} 
BN suffers from degenerate test performance if the batch data is undersized \cite{2018_ECCV_Wu}. We also show that BN suffers from the optimization difficulty with a small batch size. We show the experimental results on MNIST dataset with a batch size of 2 in Figure \ref{fig:MNIST_Experiment} (b). We find that BN can hardly learn and produces random results, while the naive network without normalization learns well. Such an observation clearly shows that BN suffers from more difficulties in training with undersized data batch.

\begin{figure}[t]
	\centering
\hspace{-0.2in}	\subfigure[]{
		\begin{minipage}[c]{.44\linewidth}
			\centering
			\includegraphics[width=4.3cm]{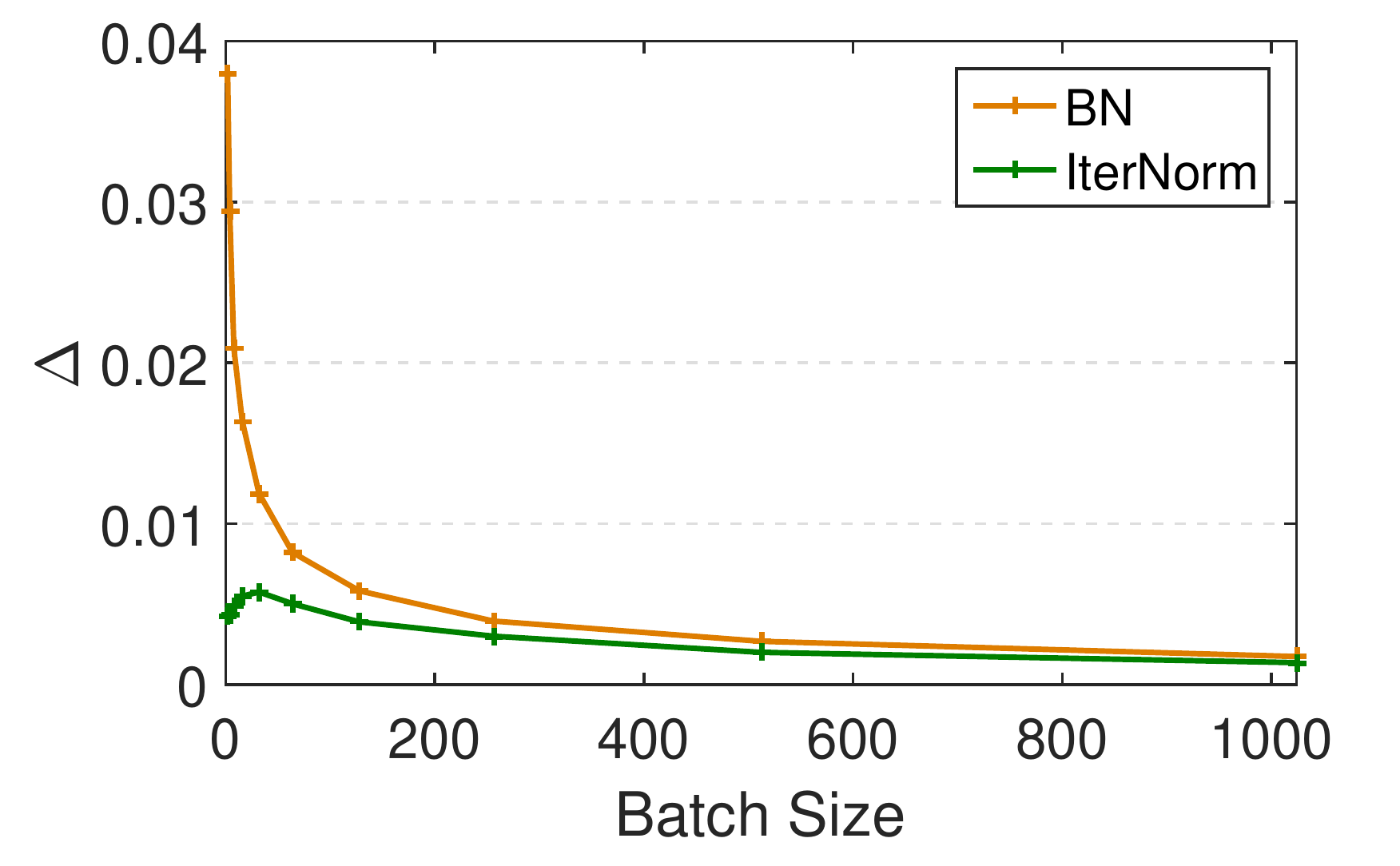}
		\end{minipage}
	}
\hspace{0.05in}	\subfigure[]{
		\begin{minipage}[c]{.44\linewidth}
			\centering
			\includegraphics[width=4.3cm]{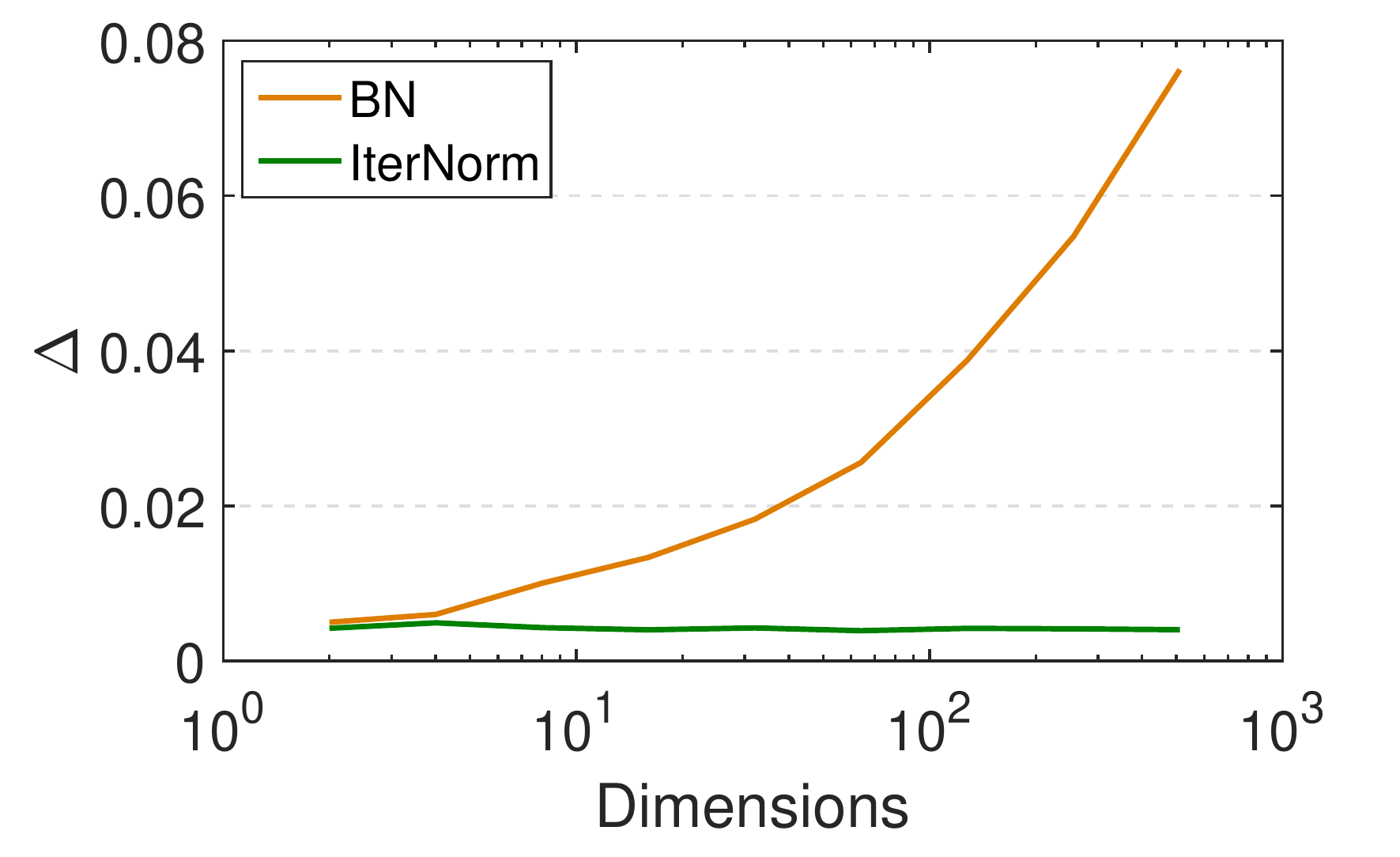}
		\end{minipage}
	}
	\caption{Illustration of the micro-batch problem of BN from the perspective of SND. (a) shows the SND with respect to batch sizes under the dimension of 128. (b) shows the SND with respect to dimensions under the batch size of 2. }
	\label{fig:micro-batch}
	\vspace{-0.2in}
\end{figure}

For an in-depth investigation, we sample the data from the dimension of 128 (Figure \ref{fig:micro-batch} (a)), and find that BN has a significantly increased SND. With increasing batch sizes, the SND of BN can be gradually reduced. Meanwhile, reduced SND leads to more stable training. 
When we fix the batch size to 2 and vary the dimension, (as shown in Figure \ref{fig:micro-batch} (b)), we observe that the SND of BN can be reduced with a low dimension. On the contrary, the SND of BN can be increased in a high-dimensional space. Thus, it can be explained why BN suffers from the difficulty during with a small batch, and why group-based normalization \cite{2018_ECCV_Wu} (by reducing the dimension and adding the examples to be standardized implicitly) alleviates the problem.

Compared to BN, IterNorm is much less sensitive to a small batch size in producing SND. Besides, the SND of 
IterNorm is more stable, even with a significantly increased dimension. Such characteristics of IterNorm are mainly attributed to its adaptive mechanism in normalization, that it stretches the dimensions along large eigenvalues and correspondingly ignores  small eigenvalues, given a fixed number of iterations \cite{2005_NumerialAlg}. 




\vspace{-0.05in}
\section{Experiments}

We evaluate IterNorm with CNNs on CIFAR datasets ~\cite{2009_TR_Alex} to show that the better optimization efficiency and generalization capability, compared to BN \cite{2015_ICML_Ioffe} and DBN \cite{2018_CVPR_Huang}. Furthermore, IterNorm with residual networks will be applied to show the performance improvement on CIFAR-10 and ImageNet \cite{2009_ImageNet} classification tasks.
The code to reproduce the experiments is available at \textcolor[rgb]{0.33,0.33,1.00}{https://github.com/huangleiBuaa/IterNorm}.

\subsection{Sensitivity Analysis}

We analyze the proposed methods on CNN architectures over the CIFAR-10 dataset \cite{2009_TR_Alex}, which contains 10 classes with $50k$ training examples and $10k$ test examples. The dataset contains $32 \times 32$ color images with 3 channels. We use the VGG networks \cite{2014_CoRR_Simonyan} tailored for $32\times 32$ inputs (16 convolution layers and 1 fully-connected layers), and the details of the networks are shown in Appendix \ref{sec:appendix_VGG}.

The datasets are preprocessed with a mean-subtraction and variance-division. We also execute normal data augmentation operation, such as a random flip and random crop with padding, as described in \cite{2015_CVPR_He}.

\begin{figure*}[t]
	\centering
	\hspace{-0.02\linewidth}
	\subfigure[basic configuration]{
		\includegraphics[width=0.25\linewidth]{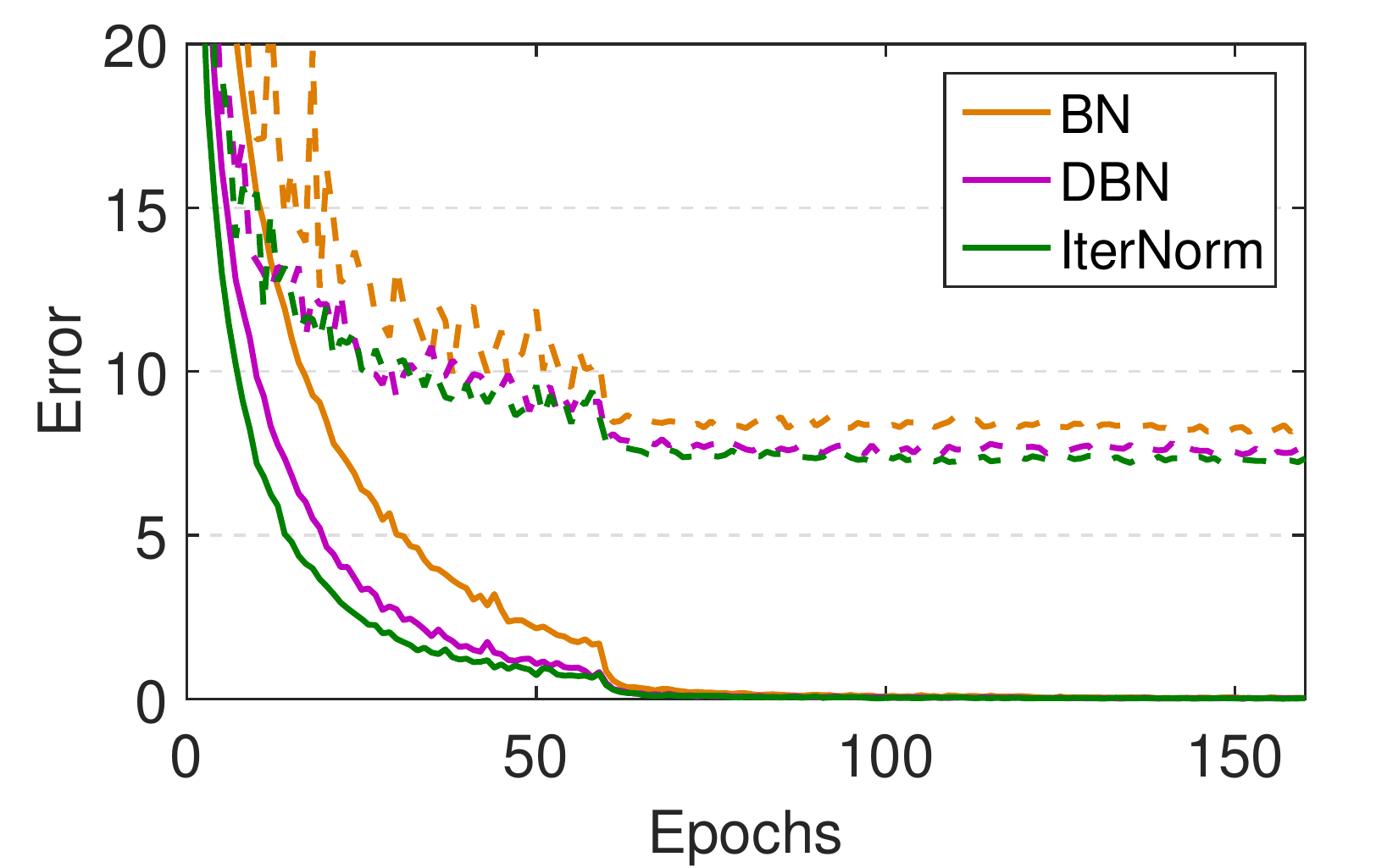}
	}
	\hspace{-0.02\linewidth}	\subfigure[batch size of 1024]{
		\includegraphics[width=0.25\linewidth]{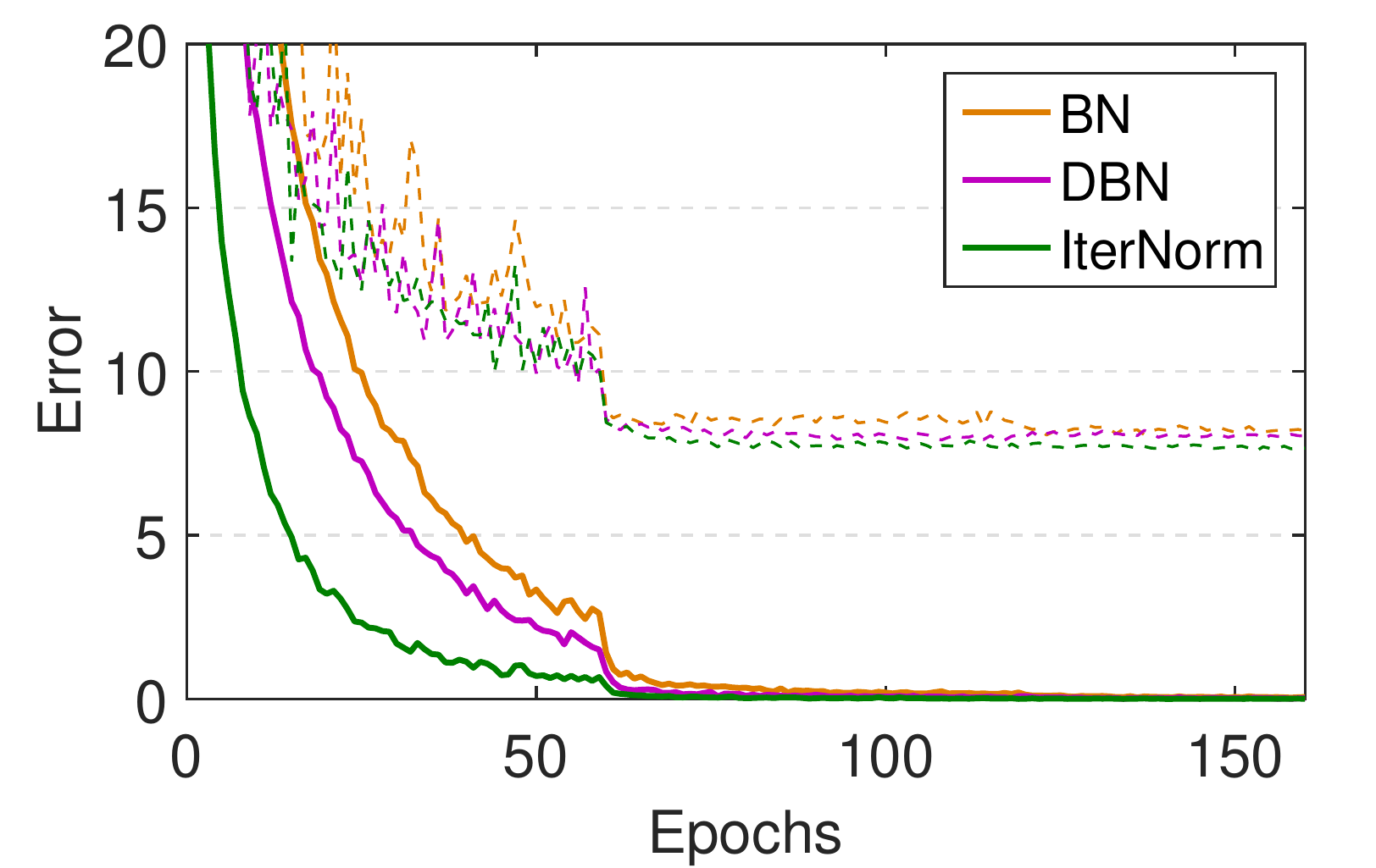}
	}
	\hspace{-0.02\linewidth}		\subfigure[batch size of 16]{
		\includegraphics[width=0.25\linewidth]{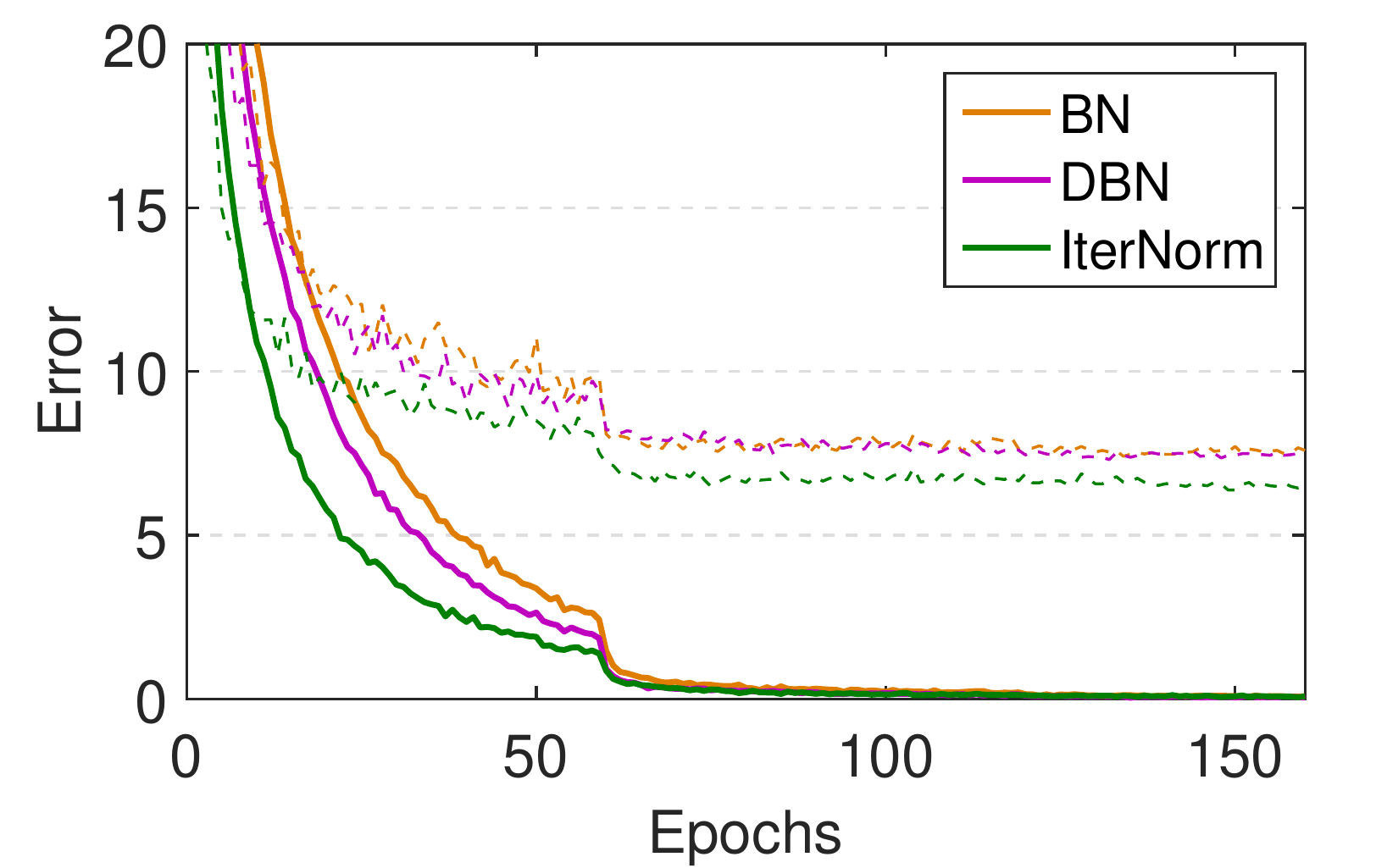}
	}
	\hspace{-0.02\linewidth}		\subfigure[10x larger learning rate]{
		\includegraphics[width=0.25\linewidth]{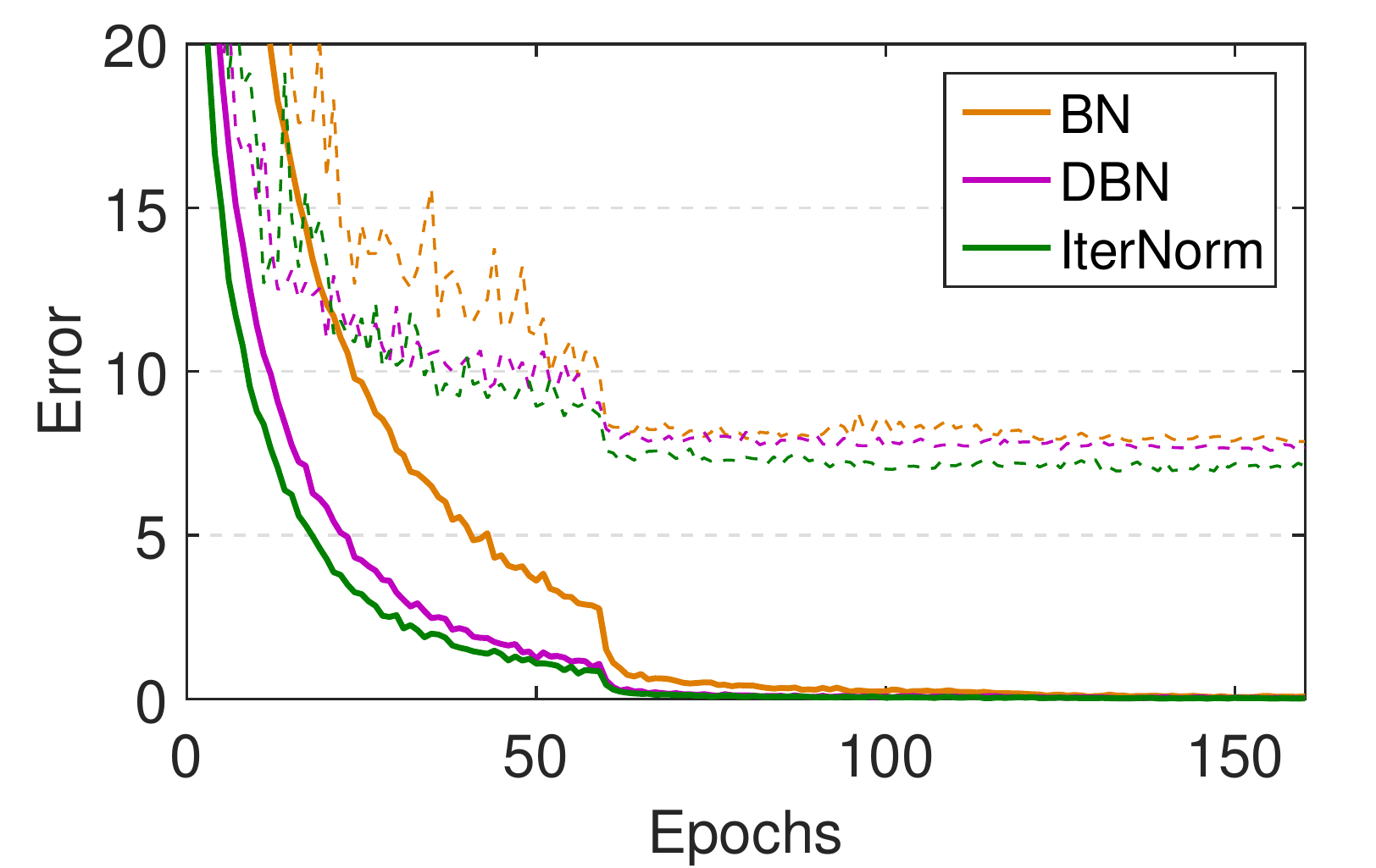}
	}
	
	\caption{\small Comparison among BN, DBN and IterNorm on VGG over CIFAR-10 datasets. We report the training (solid lines) and test (dashed lines) error with respect to epochs. }
	\label{fig:exp_Vgg}
	\vspace{-0.16in}
\end{figure*}
	\vspace{-0.12in}
\paragraph{Experimental Setup}
We use SGD with a batch size of 256 to optimize the model. We set the initial learning rate to 0.1, then divide the learning rate by 5 at  60 and 120 epochs, and finish the training at 160 epochs. All results are averaged over 3 runs.
For DBN, we use a group size of 16 as recommend in \cite{2018_CVPR_Huang}, and we find that DBN is unstable for a group size of 32 or above, due to the fact that the eigen-decomposition operation cannot converge. The main reason is that the batch size is not sufficient for DBN to full-whiten the activation for each layer.
For IterNorm, we don't use group-wise whitening in the experiments, unless otherwise stated.

\begin{figure}[t]
	\centering
	\hspace{-0.2in}	\subfigure[]{
		\begin{minipage}[c]{.44\linewidth}
			\centering
			\includegraphics[width=4.3cm]{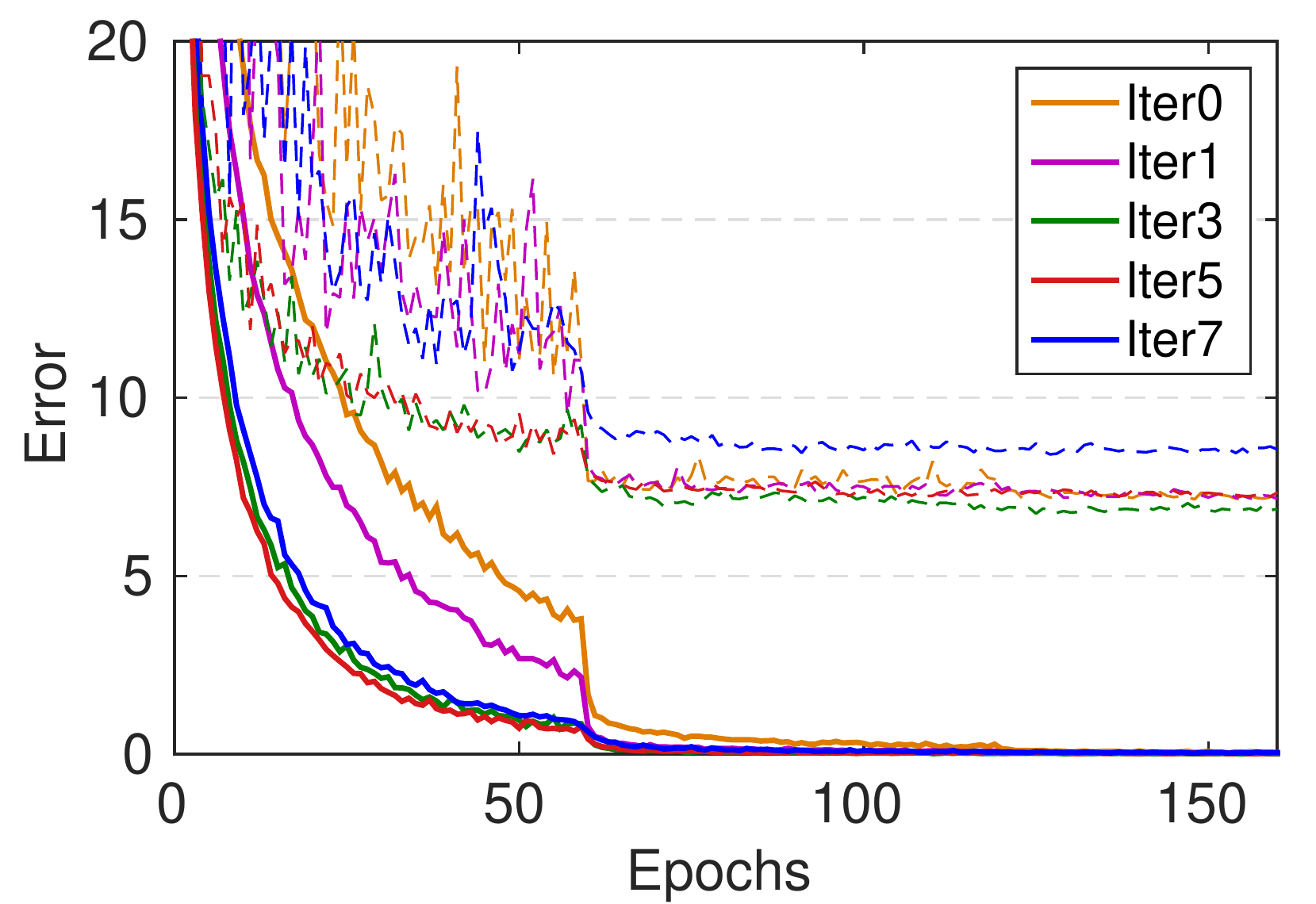}
		\end{minipage}
	}
	\hspace{0.05in}	\subfigure[]{
		\begin{minipage}[c]{.44\linewidth}
			\centering
			\includegraphics[width=4.3cm]{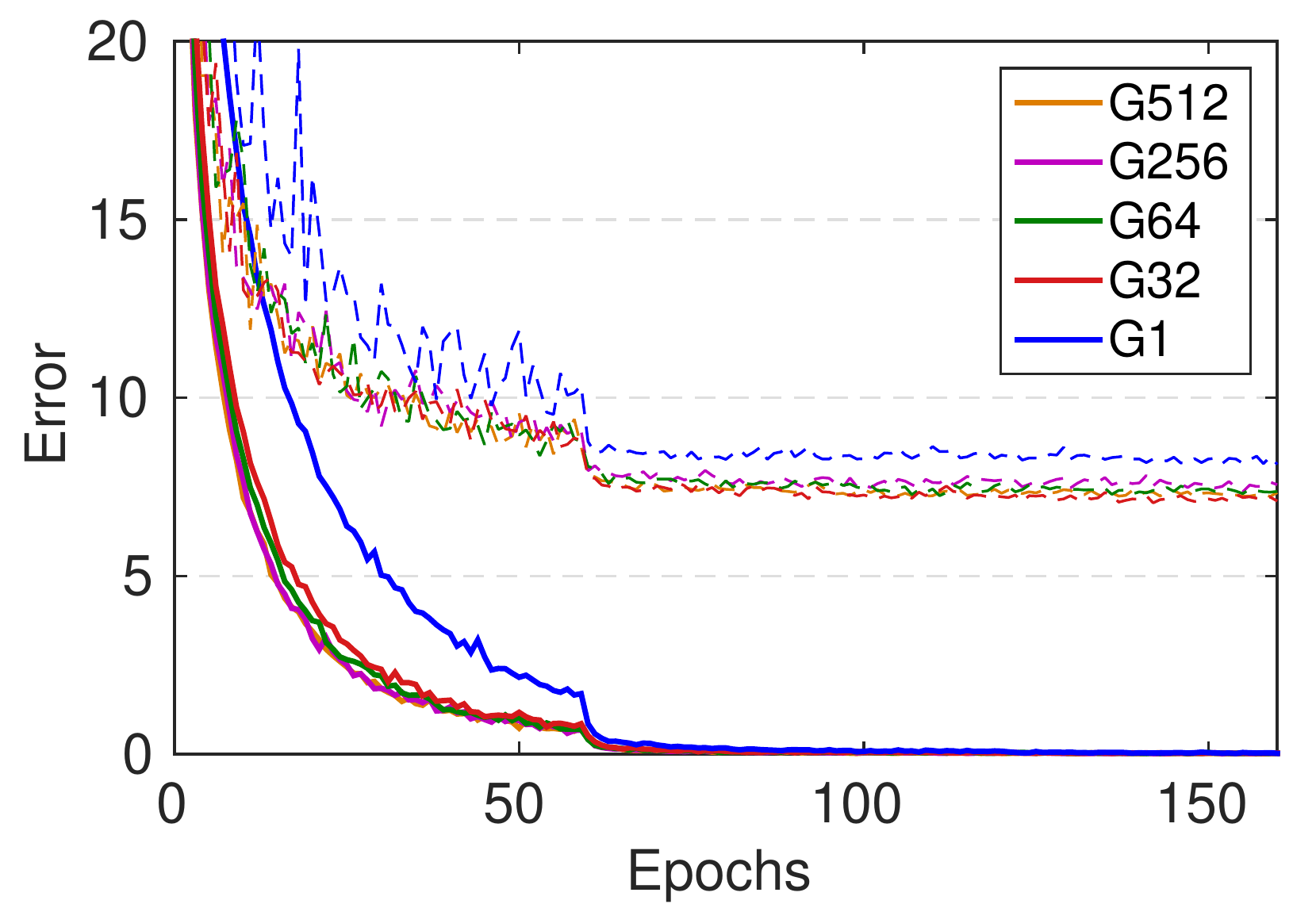}
		\end{minipage}
	}
%
	\caption{\small Ablation studies on VGG over CIFAR-10 datasets. We report the training (solid lines) and test (dashed lines) error curves. (a) shows the effects of different iteration number for IterNorm; (b) show the effects of different group size of IterNorm.}
	\label{fig:exp_Vgg_ablation}
	\vspace{-0.1in}
\end{figure}

	\vspace{-0.12in}
\paragraph{Effect of Iteration Number}
The iteration number $T$  of our IterNorm controls the extent of whitening. Here we explore the effects of $T$ on performance of IterNorm, for a range of $\{0, 1, 3, 5,7 \}$.
Note that when $T=0$, our method is reduced to normalizing the eigenvalues such that the sum of the eigenvalues is 1.
Figure \ref{fig:exp_Vgg_ablation} (a) shows the results.
We find that the smallest ($T=0$)  and the largest ($T=7$) iteration number both have the worse performance in terms of training efficiency. Further, when $T=7$, IterNorm has significantly worse test performance. These observations show that (1)
whitening within an mini-batch can improve the optimization efficiency, since IterNorm progressively stretches out the data along the dimensions of the eigenvectors such that the corresponding eigenvalue towards 1, with increasing iteration $T$; (2) controlling the extent of whitening is essential for its success, since stretching out the dimensions along small eigenvalue may produce large SND as described in Section \ref{sec_analyze_discuss}, which not only makes estimating the population statistics difficult --- therefore causing  higher test error --- but also makes optimization difficult.
 Unless otherwise stated, we use an iteration number of 5 in subsequent experiments. 

	\vspace{-0.12in}
\paragraph{Effects of Group Size}
We also investigate the effects of group size. We vary the group size in $\{ 256, 64, 32, 1 \}$, compared to the full-whitening operation of IterNorm (group size of 512). Note that our IterNorm with group size of 1, like DBN,  is also reduced to Batch Normalization \cite{2015_ICML_Ioffe}, which is ensured by Eqn. \ref{eqn:Spectral_Norm} and \ref{eqn:inv_Spectral_Norm}. 
 The results are shown in Figure \ref{fig:exp_Vgg_ablation} (b).
 We can find that our IterNorm, unlike DBN, is not sensitive to the large group size, not only in training, but also in testing. The main reason is that IterNorm  gradually stretches out the data along the dimensions of eigenvectors such that the corresponding eigenvalue towards 1, in which the speed of convergence for each dimension is proportional to the associated eigenvalues \cite{2005_NumerialAlg}. Even though there are many small eigenvalue or zero in high-dimension space, IterNorm only stretches the dimension along the associate eigenvector a little, given small iteration $T$, which introduces few SND.
In practice, we can use a smaller group size, which can reduce the computational costs. We recommend using a group size of 64, which is proposed in the experiments of Section \ref{exp_CNN} and \ref{exp_imagenet} for IterNorm.

	\vspace{-0.12in}
\paragraph{Comparison of Baselines} We compare our IterNorm with $T=5$ to BN and DBN. Under the basic configuration, we also experiment with other configurations, including (1) using a large batch size of 1024; (2) using a small batch size of 16; and (3) increasing the learning rate by 10 times and considering mini-batch based normalization is highly dependent on the batch size and their benefits comes from improved conditioning and therefore larger learning rate.  All experimental setups are the same, except that we search a different learning rate in $\{0.4, 0.1, 0.0125 \}$ for different batch sizes, based on the linear scaling rule \cite{2017_CoRR_Goyal}. Figure \ref{fig:exp_Vgg} shows the results.

We find that our proposed IterNorm converges the fastest with respect to the epochs, and generalizes the best, compared to BN and DBN. DBN also has better optimization and generalization capability than BN. Particularly, IterNorm reduces the absolute test error of BN by $0.79\%$, $0.53\%$, $1.11\%$, $0.75\%$ for the four experiments above respectively, and DBN by $0.22$, $0.37$, $1.05$, $0.58$. The results demonstrate that our IterNorm outperforms BN and DBN in terms of optimization quality and generalization capability.

\begin{figure}[t]
	\centering
	\hspace{-0.06\linewidth}	\subfigure[WRN-28-10]{
		\begin{minipage}[c]{.44\linewidth}
			\centering
			\includegraphics[width=4.2cm]{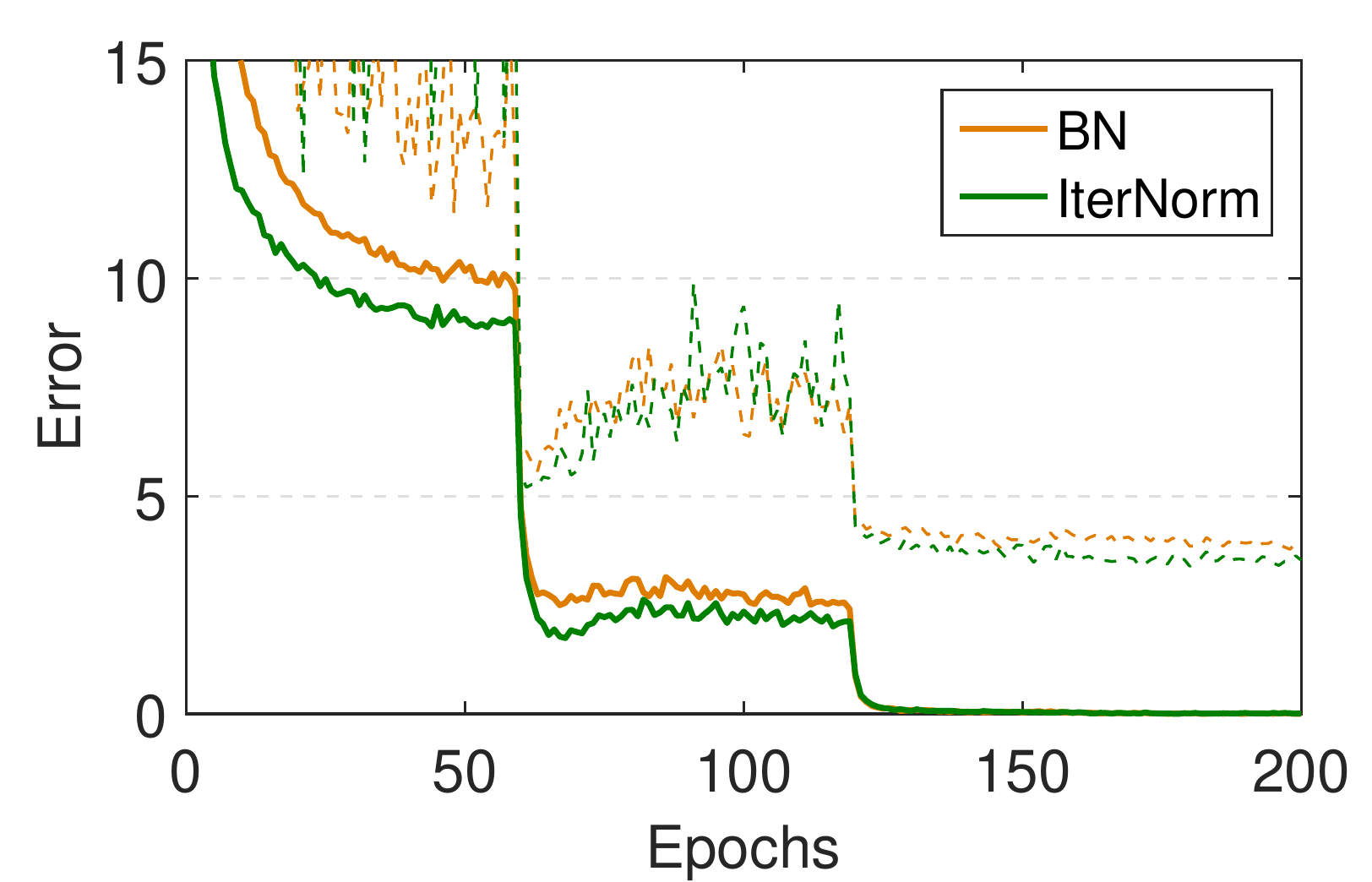}
		\end{minipage}
	}
	\hspace{0.02\linewidth}	\subfigure[WRN-40-10]{
		\begin{minipage}[c]{.44\linewidth}
			\centering
			\includegraphics[width=4.2cm]{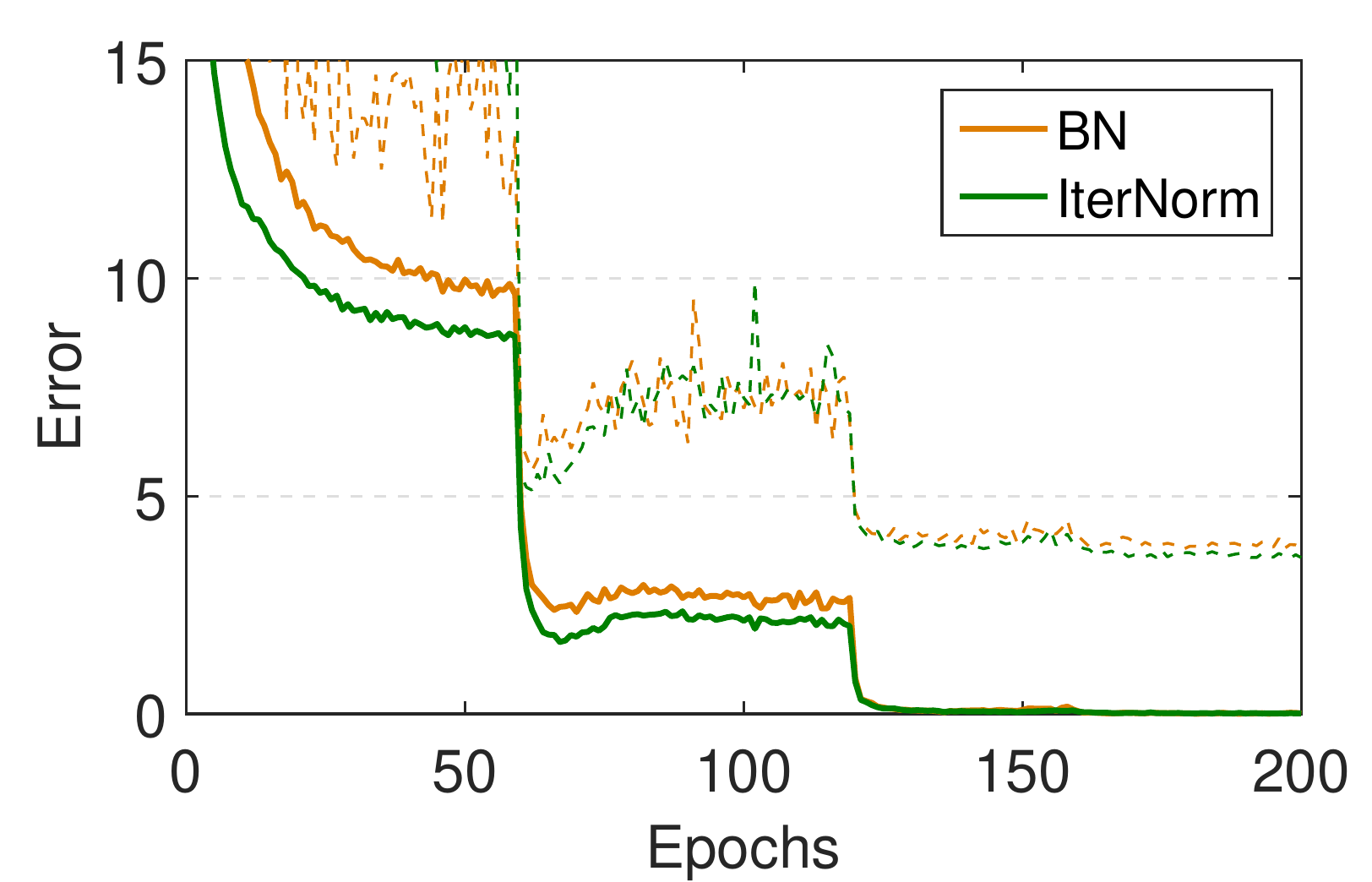}
		\end{minipage}
	}
	\caption{Comparison on Wide Residual Networks over CIFAR-10 datasets. The solid line indicates the training errors and the dashed line indicates the test errors. (a) shows the results on WRN-28-10 and (b) on WRN-40-10. }
	\label{fig:wr}
	\vspace{-0.1in}
\end{figure}
\subsection{Results on CIFAR-10 with Wide Residual Networks}
\label{exp_CNN}
We apply our IterNorm to Wide Residual Network (WRN) \cite{2016_CoRR_Zagoruyko}
to improve the performance on CIFAR-10.
Following the conventional description in \cite{2016_CoRR_Zagoruyko}, we use the abbreviation WRN-d-k to indicate a WRN with depth \emph{d} and width \emph{k}.
We  adopt the publicly available Torch implementation\footnote{https://github.com/szagoruyko/wide-residual-networks}
and follow the same setup as in \cite{2016_CoRR_Zagoruyko}.
 We apply IterNorm to WRN-28-10 and WRN-40-10 by replacing all the BN modules with our IterNorm.
 Figure \ref{fig:wr}  gives the training and testing errors with respect to the training epochs. We clearly find that the wide residual network with our proposed IterNorm improves the original one with BN, in terms of optimization efficiency and generalization capability.
 Table \ref{table:wr} shows the final test errors, compared to previously reported results for the baselines and DBN \cite{2018_CVPR_Huang}.

The results show IterNorm improves the original WRN with BN and DBN on CIFAR-10.
In particular, our methods reduce the test error to $3.56 \%$ on WRN-28-10, a relatively improvement of $8.5 \%$ in performance over `Baseline'.

\begin{table}[]
	\centering
	\begin{small}
		\begin{tabular}{c|cc}
			\toprule
			Method & WRN-28-10 & WRN-40-10     \\
			\hline
			 Baseline*~\cite{2016_CoRR_Zagoruyko}  & 3.89  & 3.80    \\
			 DBN ~\cite{2018_CVPR_Huang}  & 3.79  $\pm$ 0.09  & 3.74  $\pm$ 0.11   \\
			 Baseline  & 3.89 $\pm$ 0.13  & 3.82 $\pm$ 0.11   \\
			 IterNorm  & \textbf{ 3.56 $\pm$ 0.12}  & \textbf{3.59 $\pm$ 0.07}  \\
			\bottomrule
		\end{tabular}
		\vspace{0.1in}
		\caption{Test errors ($\%$) on wide residual networks over CIFAR-10. All results are computed over 5 random seeds, and shown in the format
			of `mean $\pm std$'. We replicate the  `Baseline' results based on the released code in
			\cite{2016_CoRR_Zagoruyko}, which computes the median of 5 runs on WRN-28-10 and only performs one run onWRN-40-10.}
		\label{table:wr}
		
	\end{small}
\end{table}

\begin{table}[t]
	\centering
	\begin{small}
		\begin{tabular}{c|cc}
			\toprule
			Method     & Top-1   & Top-5 \\
			\hline
			Baseline* \cite{2015_CVPR_He} &  30.43 & 10.76  \\
			DBN-L1* \cite{2018_CVPR_Huang} & 29.87 & 10.36  \\
			Baseline     & 29.76 & 10.39  \\
			DBN-L1   & 29.50 & 10.26  \\
			IterNorm-L1     & 29.34 & 10.22  \\		
			IterNorm-Full     & 29.30 & 10.21 \\
			IterNorm-L1~~+ ~~DF   &\textbf{28.86} & \textbf{10.08} \\
			\bottomrule
		\end{tabular}
		\vspace{0.1in}
		\caption{Comparison of validation errors ($\%$, single model and
			single-crop) on 18-layer residual networks
			on ILSVRC-2012.\hspace{\textwidth} `Baseline*' and `DBN-L1*' indicate that the results are reported in \cite{2018_CVPR_Huang} with training of 90 epochs.}
		\label{table:ImageNet-Res-18}
	\end{small}
\end{table}

\subsection{Results on ImageNet with Residual Network}
\label{exp_imagenet}

We validate the effectiveness of our methods on residual networks for ImageNet classification with 1000 classes ~\cite{2009_ImageNet}.
We use the given official 1.28M training images as a training set, and evaluate the top-1 and top-5 classification errors on the validation
set with 50k images.
	\vspace{-0.12in}
\paragraph{Ablation Study on Res-18} We first execute an ablation study on the 18-layer residual network (Res-18) to explore multiple positions for replacing BN with IterNorm. The models used are as follows: (a) `IterNorm-L1': we only replace the first BN module of ResNet-18, so that the decorrelated information from previous layers can pass directly to the later layers with the identity connections described in \cite{2018_CVPR_Huang}; (b) We also replace all BN modules indicated as `IterNorm-full';
We follow the same experimental setup as described in ~\cite{2015_CVPR_He}, except that we use 1 GPU and train over 100 epochs. We apply SGD with a mini-batch size of 256, momentum of 0.9 and weight decay of 0.0001. The initial learning rate is set to 0.1 and divided by 10 at 30, 60  and 90 epochs, and end the training at 100 epochs.

We find that only replacing the first BN effectively improves the performance of the original residual network, either by using DBN or IterNorm.
Our IterNorm has marginally better performance than DBN.
 We find that replacing all the layers of IterNorm has no significant improvement over only replacing the first layer. We conjecture that the reason might be that the learned residual functions tend to have small response as shown in \cite{2015_CVPR_He}, and stretching this small response to the magnitude as the previous one may lead to negative effects. Based on `IterNorm-L1', we further plug-in the IterNorm after the last average pooling (before the last linear layer) to learn the decorrelated feature representation. We find this significantly improves the performance, as shown in Table \ref{table:ImageNet-Res-18}, referred to as `IterNorm-L1 + DF'. Such a way to apply IterNorm can improve the original residual networks and introduce negligible computational cost.
  We also attempt to use DBN to decorrelate the feature representation. However, it always suffers the problems of that the eigen-decomposition can not converge.

	\vspace{-0.12in}
\paragraph{Results on Res-50/101}
We further apply our method on the 50 and 101-layer residual network (ResNet-50 and ResNet-101)
and perform single model and single-crop testing.
We use the same experimental setup as before, except that we use 4 GPUs and train over 100 epochs. The results are shown in Table
\ref{table:ImageNet}. We can see that the `IterNorm-L1' achieves lower
test errors compared to the original residual networks. `IterNorm-L1~ +~ DF ' further improves the performance.
\begin{table}[t]
	\centering
	\begin{small}
		\begin{tabular}{c|cc|cc}
			\toprule
			& \multicolumn{2}{c|}{Res-50} &    \multicolumn{2}{c}{Res-101}   \\
			Method   & Top-1    & Top-5  & Top-1   & Top-5 \\
			\hline
			Baseline* \cite{2015_CVPR_He}  & 24.70  &  7.80   & 23.60 & 7.10  \\
			Baseline    & 23.95 &  7.02   & 22.45 & 6.29  \\
			IterNorm-L1    & 23.28 &  6.72   & 21.95 & 5.99  \\
			IterNorm-L1~~+ ~~DF  &\textbf{22.91}&\textbf{6.47}&\textbf{21.77} & \textbf{5.94} \\
			\bottomrule
		\end{tabular}
		\vspace{0.1in}
		\caption{Comparison of test errors ($\%$, single model and
			single-crop) on 50/101-layer residual networks
			on ILSVRC-2012.\hspace{\textwidth} `Baseline*' indicates that the results are obtained from the website: https://github.com/KaimingHe/deep-residual-networks.}
		\label{table:ImageNet}
	\end{small}
	\vspace{-0.12in}
\end{table}


\section{Conclusions}

In this paper, we proposed Iterative Normalization (IterNorm) based on Newton's iterations. It improved the optimization efficiency and generalization capability over standard BN by decorrelating activations, and improved the efficiency over DBN by avoiding the computationally expensive eigen-decomposition.
 We introduced Stochastic Normalization Disturbance (SND) to measure the inherent stochastic uncertainty in normalization. With the support of SND,  we provided a thorough analysis regarding the performance of normalization methods, with respect to the batch size and feature dimensions, and showed that IterNorm has better trade-off between optimization and generalization. We demonstrated consistent performance improvements of IterNorm on the CIFAR-10 and ImageNet datasets.
The analysis of combining conditioning and SND, can potentially lead to novel visions for future normalization work, and our proposed IterNorm can potentially  to be used in designing network architectures. 




\appendix

\renewcommand{\thetable}{A\arabic{table}}
\setcounter{table}{0}

\renewcommand{\thefigure}{A\arabic{figure}}
\setcounter{figure}{0}

\section{Derivation of Back-propagation}
\label{sec:appendix_back}
We first show the forward pass for illustration. We follow the common matrix notation that the vectors are column vectors by default while their derivations are row  vectors.  Given the mini-batch inputs  $ \mathbf{X} \in \mathbb{R}^{d \times m} $, the forward pass of our Iterative Normalization (IterNorm) to compute the whitened output is described below:
\begin{eqnarray}
\label{eqn:mu} 
\mathbf{\mu} = \frac{1}{m} \mathbf{X} \cdot \mathbf{1}  \\
\label{eqn:XC} 
\mathbf{X}_C = \mathbf{X}-\mathbf{\mu}  \cdot \mathbf{1}^T \\
\label{eqn:sigma} 
\Sigma = \frac{1}{m}\mathbf{X}_C \mathbf{X}_C ^T + \epsilon \mathbf{I}\\
\label{eqn:sigmaN} 
\Sigma_{N} = \Sigma / tr(\Sigma)\\
\label{eqn:sigmaSquare} 
\Sigma^{-\frac{1}{2}} = \mathbf{P}_T / \sqrt{tr(\Sigma)}\\
\label{eqn:Xhat} 
\widehat{\mathbf{X}} = \Sigma^{-\frac{1}{2}} \mathbf{X}_C 
\end{eqnarray}
where $\mathbf{P}_T$ is calculated based on Newton's iterations as follows:
\begin{equation}
\label{eqn:Iteration}
\begin{cases}
\mathbf{P}_0=\mathbf{I} \\
\mathbf{P}_{k}=\frac{1}{2} (3 \mathbf{P}_{k-1} - \mathbf{P}_{k-1}^{3} \Sigma_{N}), ~~ k=1,2,...,T.
\end{cases}
\end{equation}

The back-propagation pass is based on the chain rule. 
Given $ \frac{\partial L}{\partial \widehat{\mathbf{X}}}$, we can calculate $ \frac{\partial L}{\partial \mathbf{X}}$ based on Eqn. \ref{eqn:mu} and \ref{eqn:XC}:
\begin{equation}
\frac{\partial L}{\partial \mathbf{X}} = \frac{1}{m} \cdot (\mathbf{1} \cdot \frac{\partial L}{\partial \mathbf{\mu}})^T +  \frac{\partial L}{\partial \mathbf{X}_C},  
\end{equation}
where $\frac{\partial L}{\partial \mathbf{\mu}}$ is calculated based on Eqn. \ref{eqn:XC}:
\begin{eqnarray}
\frac{\partial L}{\partial \mathbf{\mu}} = - \mathbf{1}^T \cdot \frac{\partial L}{\partial \mathbf{X}_C}^T    
\end{eqnarray}
and $ \frac{\partial L}{\partial \mathbf{X}_C}$ is calculated based on Eqn. \ref{eqn:sigma} and \ref{eqn:Xhat}:  
\begin{eqnarray}
\frac{\partial L}{\partial \mathbf{X}_C}  = \ \Sigma^{-\frac{1}{2}} \frac{\partial L}{\partial \widehat{\mathbf{X}}}  + \frac{2}{m}  (\frac{\partial L}{\partial \Sigma})_{s} \mathbf{X}_C,
\end{eqnarray}
where $ (\frac{\partial L}{\partial \Sigma})_{s}$ means symmetrizing $\frac{\partial L}{\partial \Sigma}$ by $\frac{\partial L}{\partial \Sigma}= \frac{1}{2}(\frac{\partial L}{\partial \Sigma} + \frac{\partial L}{\partial \Sigma}^T)$.  Next, we will calculate $\frac{\partial L}{\partial \Sigma}$, given $\frac{\partial L}{\partial \Sigma^{-\frac{1}{2}}}$ as follows:
\begin{eqnarray}
\frac{\partial L}{\partial \Sigma^{-\frac{1}{2}}}	 =\frac{\partial L}{\partial \widehat{\mathbf{X}}}     \mathbf{X}_C^T
\end{eqnarray}

\begin{table}[t]
	\centering
	\begin{small}
		\begin{tabular}{c|p{1.8cm}<{\centering}p{1.8cm}<{\centering}}
			\toprule
			Module  & d=64 & d=128     \\
			\hline
			`nn' convolution  & 8.89ms  & 17.46ms    \\
			`cudnn' convolution  & 5.65ms    & 13.62ms    \\
			DBN \cite{2018_CVPR_Huang}  & 15.92ms   & 35.02ms \\
			IterNorm-iter3  & 6.59ms   & 13.32ms    \\
			IterNorm-iter5  & 7.40ms   & 13.92ms \\
			IterNorm-iter7  &  8.21ms   & 14.68ms   \\
			\bottomrule
		\end{tabular}
		\vspace{0.1in}
		\caption{Comparison of wall clock time (ms). `IterNorm-iterN' (N=3, 5 and 7) indicates the proposed IterNorm with iteration number of N. We compute the time including the forward pass and the backward pass, averaged over 100 runs. }
		\label{table:timeCost}
		
	\end{small}
\end{table}

Based on Eqn. \ref{eqn:sigmaN} and \ref{eqn:sigmaSquare}, we obtain:
\begin{small}
	\begin{align}
	\frac{\partial{L}}{\partial{\Sigma}} &=   \frac{\partial{L}}{\partial{\Sigma_{N}}}    \frac{\Sigma_{N}}{\partial{\Sigma}}  
	+  \frac{\partial{L}}{\partial{\Sigma^{-\frac{1}{2}}}}   \frac{\partial \Sigma^{-\frac{1}{2}}}{\partial{\Sigma}}  \nonumber  \\
	&=  \frac{1}{tr(\Sigma)} \frac{\partial{L}}{\partial{\Sigma_{N}}}
	+ tr(\frac{\partial{L}}{\partial{ \Sigma_{N}}}^T \Sigma)  \frac{\partial{\frac{1}{tr(\Sigma)}}}{\partial{tr(\Sigma)}}  \frac{\partial{tr(\Sigma)}}{\partial \Sigma}  \nonumber \\
	~~~&\phantom{{}={}}+ tr(\frac{\partial{L}}{\partial{ \Sigma^{-\frac{1}{2}}}}^T \mathbf{P}_T)  \frac{\partial{\frac{1}{\sqrt{tr(\Sigma)}}}}{\partial{tr(\Sigma)}}  \frac{\partial{tr(\Sigma)}}{\partial \Sigma}   \nonumber \\
	&=  \frac{1}{tr(\Sigma)} \frac{\partial{L}}{\partial{\Sigma_{N}}}   
	-\frac{1}{(tr(\Sigma))^2} tr(\frac{\partial{L}}{\partial{ \Sigma_{N}}}^T \Sigma)   \mathbf{I}   \nonumber \\
	~~~&\phantom{{}={}}- \frac{1}{2(tr(\Sigma))^{3/2}} tr((\frac{\partial{L}}{\partial{\Sigma^{-1/2}}})^T \mathbf{P}_T)  \mathbf{I}.
	\end{align}
\end{small}

Based on Eqn. \ref{eqn:Iteration}, we can derive $\frac{\partial{L}}{\partial{\Sigma_{N}}}$ as follows:
\begin{small}
	\begin{eqnarray}
	\label{eqn:backward-1}
	\frac{\partial{L}}{\partial{\Sigma_{N}}}&=& -\frac{1}{2} \sum_{k=1}^{T} (\mathbf{P}_{k-1}^3)^T \frac{\partial{L}}{\partial{\mathbf{P}_k}}  
	\end{eqnarray}
\end{small}
where $\{\frac{\partial{L}}{\partial{\mathbf{P}_{k}}},k=1,...,T \}$ can be calculated based on Eqn. \ref{eqn:sigmaSquare} and \ref{eqn:Iteration} by following iterations: 
\begin{small}
	\begin{equation}
	\label{eqn:backIteration}
	\begin{cases}
	\frac{\partial{L}}{\partial{\mathbf{P}_{T}}}= \frac{1}{\sqrt{tr(\Sigma)}} \frac{\partial{L}}{\partial{\Sigma^{-\frac{1}{2}}}}   \\
	\frac{\partial{L}}{\partial{\mathbf{P}_{k-1}}} =\frac{3}{2} \frac{\partial{L}}{\partial{\mathbf{P}_{k}}}
	-\frac{1}{2} \frac{\partial{L}}{\partial{\mathbf{P}_{k}}}  (\mathbf{P}_{k-1}^2 \Sigma_{N})^T
	-\frac{1}{2}  (\mathbf{P}_{k-1}^2)^T  \frac{\partial{L}}{\partial{\mathbf{P}_{k}}} \Sigma_{N}^T
	\\
	-  \frac{1}{2}(\mathbf{P}_{k-1})^T \frac{\partial{L}}{\partial{\mathbf{P}_{k}}} (\mathbf{P}_{k-1} \Sigma_{N})^T
	,  ~~k=T,...,1.
	\end{cases}
	\end{equation}
\end{small}

Further, we can simplify the derivation of $ \frac{\partial L}{\partial \mathbf{X}}$ as:
\begin{small}
	\begin{align}
	\frac{\partial L}{\partial \mathbf{X}} &= \frac{1}{m} \cdot (\mathbf{1} \cdot \frac{\partial L}{\partial \mathbf{\mu}})^T +  \frac{\partial L}{\partial \mathbf{X}_C}   \nonumber \\
	&=  \frac{\partial L}{\partial \mathbf{X}_C} (\mathbf{I} - \frac{1}{m} \mathbf{1} \mathbf{1}^T) \nonumber \\
	&=(\ \Sigma^{-\frac{1}{2}} \frac{\partial L}{\partial \widehat{\mathbf{X}}}  + \frac{2}{m}  (\frac{\partial L}{\partial \Sigma})_{s} \mathbf{X}_C) (\mathbf{I} - \frac{1}{m} \mathbf{1} \mathbf{1}^T)  \nonumber \\
	&= (\ \Sigma^{-\frac{1}{2}} \frac{\partial L}{\partial \widehat{\mathbf{X}}})   (\mathbf{I} - \frac{1}{m} \mathbf{1} \mathbf{1}^T) 
	+ \frac{2}{m}  (\frac{\partial L}{\partial \Sigma})_{s} \mathbf{X}_C \nonumber \\
	&\phantom{{}={}} - \frac{2}{m} \frac{1}{m} (\frac{\partial L}{\partial \Sigma})_{s} (\mathbf{X}_C \mathbf{1}) \mathbf{1}^T  \nonumber\\
	&= \ \Sigma^{-\frac{1}{2}} \frac{\partial L}{\partial \widehat{\mathbf{X}}}   (\mathbf{I} - \frac{1}{m} \mathbf{1} \mathbf{1}^T) 
	+ \frac{2}{m}  (\frac{\partial L}{\partial \Sigma})_{s} \mathbf{X}_C +0  \nonumber  \\
	&= \ \Sigma^{-\frac{1}{2}}   ( \frac{\partial L}{\partial \widehat{\mathbf{X}}} - \mathbf{f} \mathbf{1}^T) 
	+ \frac{2}{m}  (\frac{\partial L}{\partial \Sigma})_{s} \mathbf{X}_C   \nonumber \\
	&= \ \Sigma^{-\frac{1}{2}}   ( \frac{\partial L}{\partial \widehat{\mathbf{X}}} - \mathbf{f} \mathbf{1}^T) 
	+ \frac{1}{m}  (\frac{\partial L}{\partial \Sigma}+ \frac{\partial L}{\partial \Sigma}^T) \mathbf{X}_C
	\end{align}
\end{small}
where $\mathbf{f}=\frac{1}{m} \frac{\partial L}{\partial \widehat{\mathbf{X}}} \cdot \mathbf{1} $.

\section{Comparison of Wall Clock Time }
\label{sec:appendix_time}

As discussed in Section 3.3 of the main paper, the computational cost of our method is comparable to the convolution operation. To be specific, given the internal activation $\mathbf{X}_C \in \mathbb{R}^{h \times w \times d  	\times m} $, the $3 \times 3$ convolution with the same input and output feature maps costs $9hwmd^2$, while our IterNorm costs $2hwmd^2 + T d^3$. The relative cost of IterNorm for $3 \times 3$ convolution is $2/9 + Td/mhw$. 

Here, we compare the wall-clock time of IterNorm, Decorrelated Batch Normalization (DBN) \cite{2018_CVPR_Huang} and $3 \times 3$ convolution.
Our IterNorm is implemented based on Torch \cite{2011_torch}. 
The implementation of DBN is from the released code of the DBN paper \cite{2018_CVPR_Huang}. 
We compare `IterNorm' to the `nn', `cudnn' convolution \cite{2014_cudnn} and DBN \cite{2018_CVPR_Huang} in Torch. The experiments are run on a TITAN Xp. We use the  corresponding configurations of the input $\mathbf{X}_C \in \mathbb{R}^{h \times w \times d  	\times m} $  and convolution $\mathbf{W}_C \in \mathbb{R}^{3 \times 3 \times d  	\times d}$: $m=64$, $h=w=32$. We compare the results of $d=64$ and $d=128$, as shown in Table \ref{table:timeCost}. We find that our unoptimized implementation of IterNorm (\eg, `IterNorm-iter5') is faster than the `nn' convolution, and slightly slower than `cudnn' convolution. Note that our IterNorm is implemented based on the API provided by Torch \cite{2011_torch}, it is thus more fair to compare IterNorm to `nn' convolution.
Besides, our IterNorm is significantly faster than DBN. 

Besides, we also conduct additional experiments to compare the training time of IterNorm and DBN on the VGG architecture described in Section 5.1 of the main paper, with a batch size of 256. 
DBN (group size of 16) costs 1.366s per iteration, while IterNorm costs 0.343s.

\begin{figure}[t]
	\centering
	\hspace{-0.1in}	\subfigure[]{
		\centering
		\includegraphics[width=4.0cm]{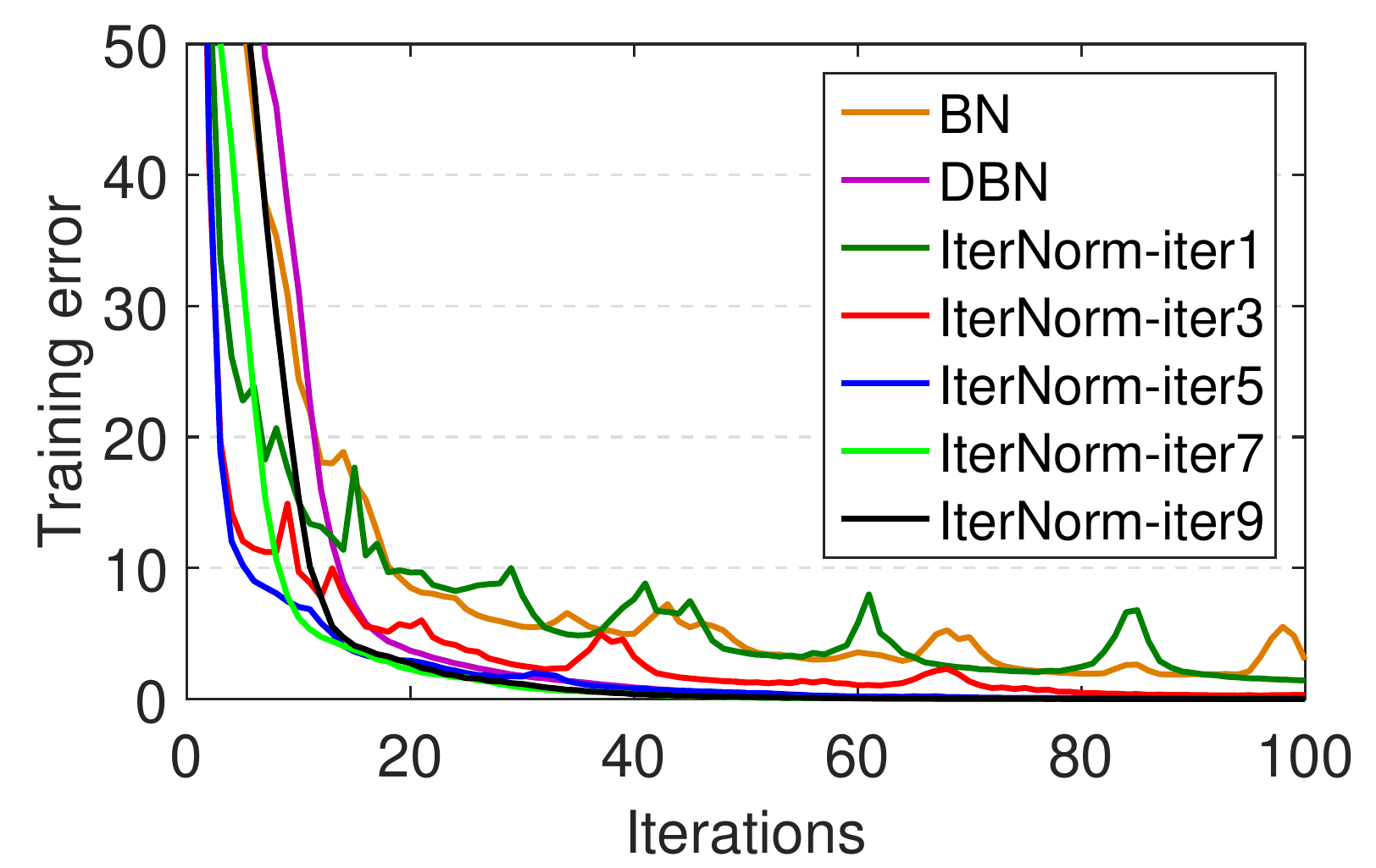}
	}
	\hspace{-0.1in}	\subfigure[]{
		\centering
		\includegraphics[width=4.0cm]{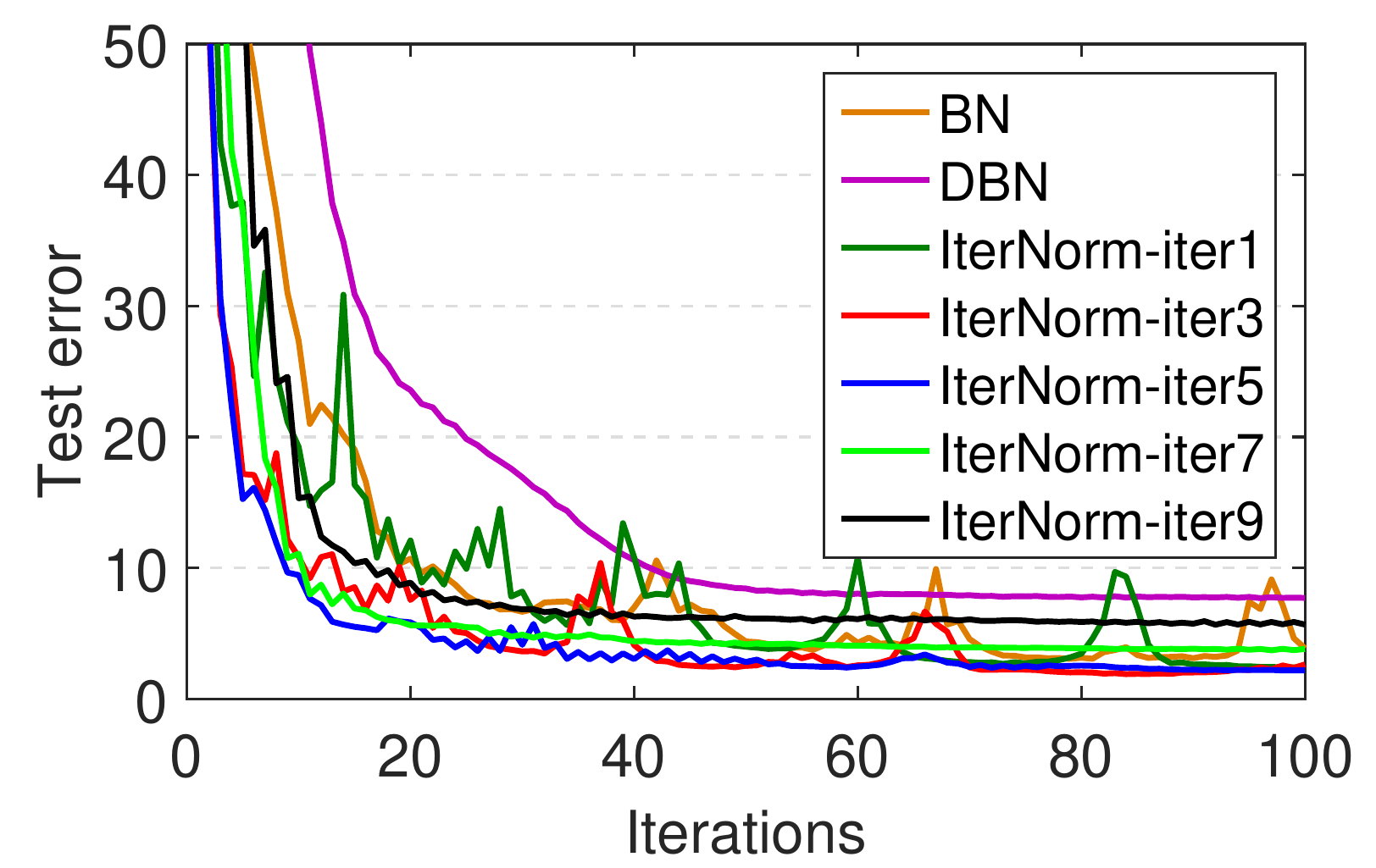}
	}	
	\caption{\small Ablation study in training an MLP on MNIST. The experimental setup is the same as the setup in Section 4.2 of the main paper: We train a 4-layer MLP and the number of neurons in each hidden layer is 100; We use full batch gradient and report the best results with respect to the training loss among learning rates=$\{0.2,0.5,1,2,5\}$. (a) shows the training error with respect to the iterations, and (b)  shows the test error with respect to the iterations.}
	\label{fig:exp_MLP}
		\vspace{-0.2in}
\end{figure}

\begin{figure}[t]
	\centering
	\hspace{-0.1in}	\subfigure[]{
		\centering
		\includegraphics[width=4.0cm]{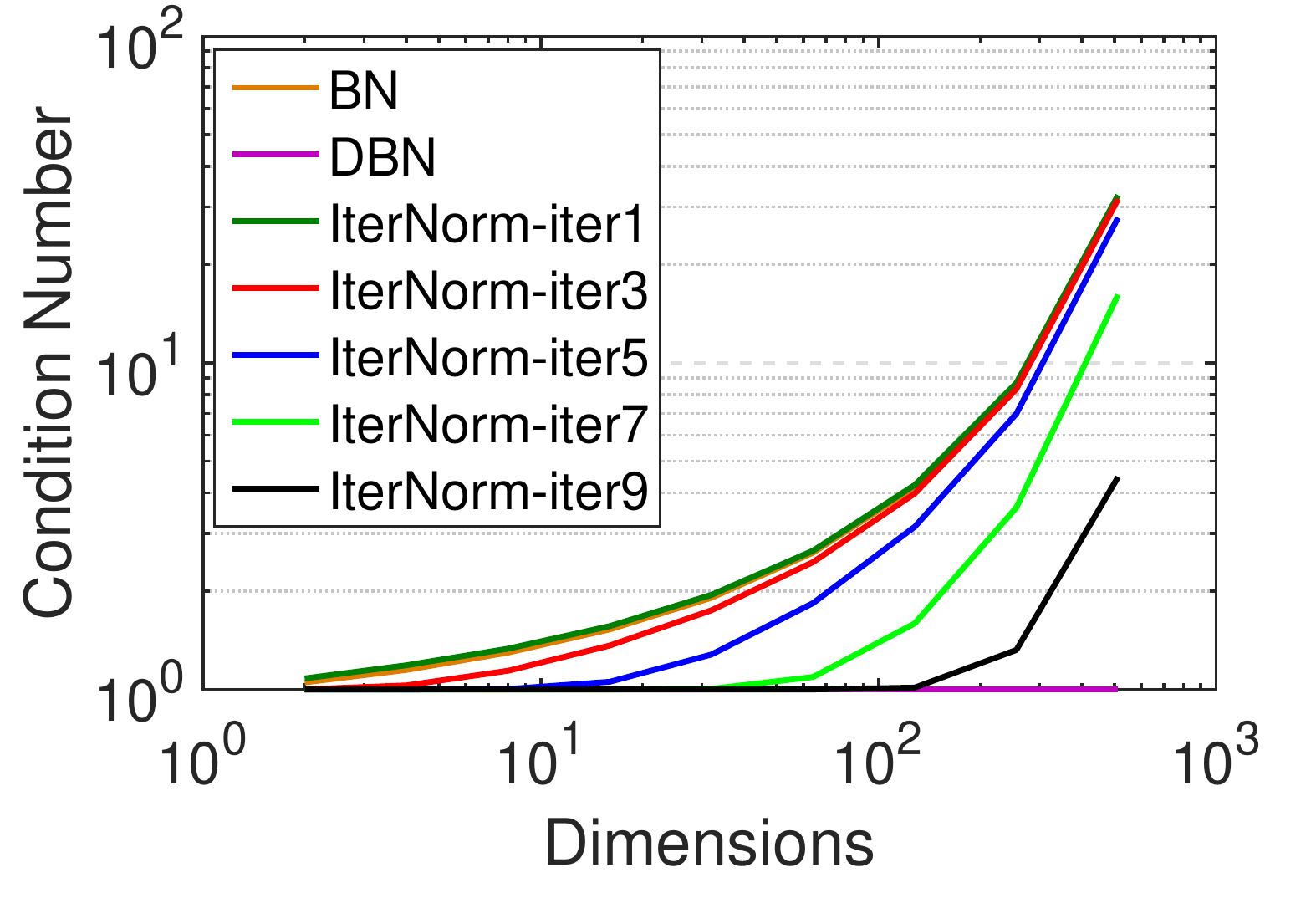}
	}
	\hspace{-0.1in}	\subfigure[]{
		\centering
		\includegraphics[width=4.0cm]{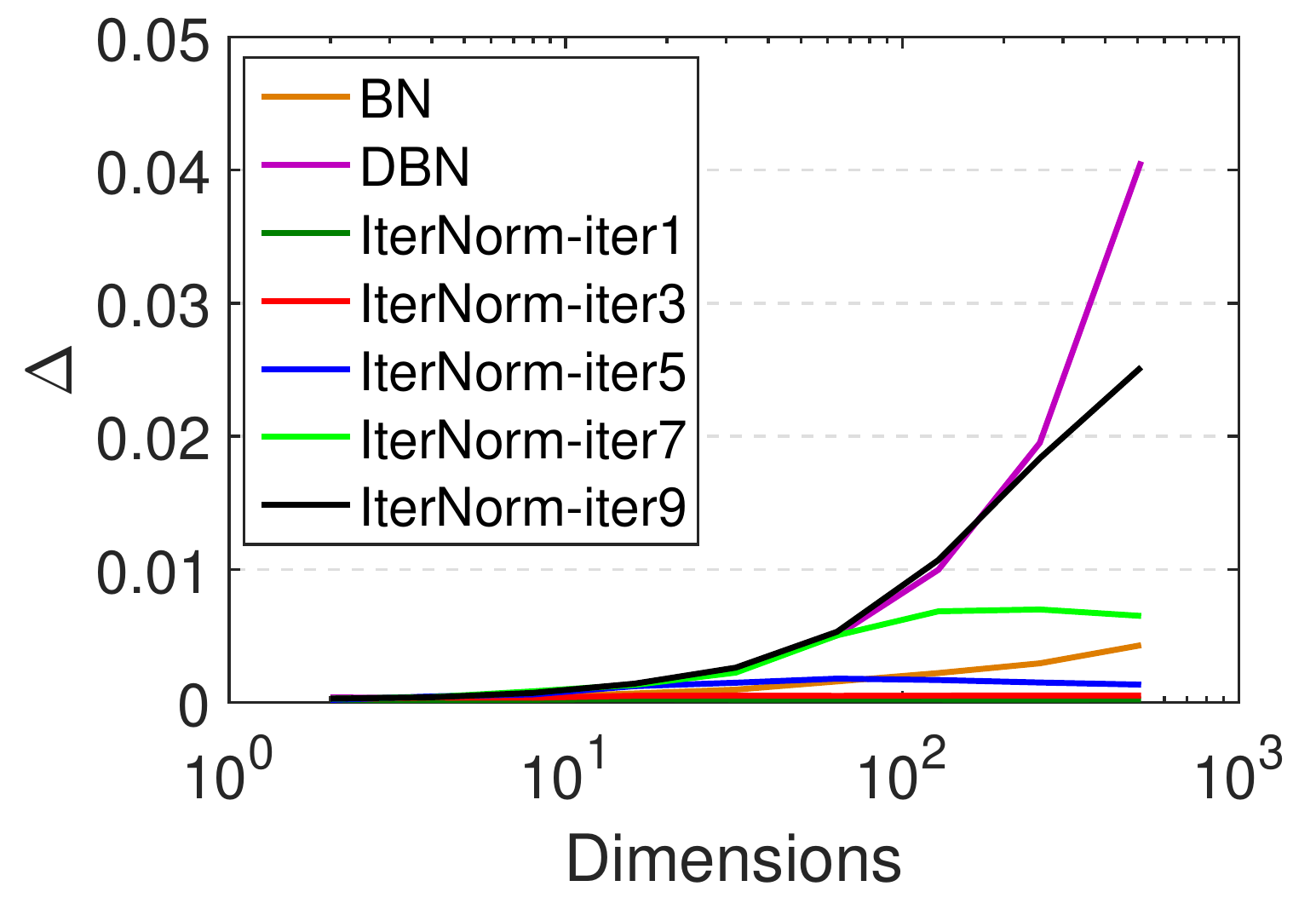}
	}	
	\caption{\small Comparison of different normalization operations in condition number of covariance matrix of normalized output (a) and SND (b). We sample 60,000 examples from Gaussian distribution and choose a batch size of $1024$, and observe the results with respect to the dimensions from $2^1$ to $2^9$, averaged over 10 times.}
	\label{fig:exp_SND}
\end{figure}

\begin{table}[]
	\centering
	\begin{small}
		\begin{tabular}{|p{3.8cm}<{\centering}|}
			\hline
			input ($32 \times 32$ RGB image)     \\
			\hline
			conv3(3,64)    \\
			conv3(64,64)    \\
			maxpool(2,2) \\
			\hline
			conv3(64,128)    \\
			conv3(128,128)    \\
			maxpool(2,2) \\		
			\hline
			conv3(128,256)    \\
			conv3(256,256) $\times 3$    \\
			maxpool(2,2) \\		
			\hline
			conv3(256,512)    \\
			conv3(512,512)  $\times 3$   \\
			maxpool(2,2) \\		
			\hline
			conv3(512,512) $\times 4$   \\
			avepool(2,2) \\			
			\hline	
			FC(512,10) \\
			\hline	
			soft-max \\			
			\hline
		\end{tabular}
		\vspace{0.1in}
		\caption{The VGG network used in the experiment as shown in Section 5.1 in the main paper. `conv3($d_{in}$, $d_{out}$)' indicates the $3 \times 3$ convolution with input channel number of $d_{in}$ and  output channel number of $d_{out}$.}
		\label{table:VGG}
		
	\end{small}
\end{table}

\section{Experiments of IterNorm with Different Iterations}

\label{sec:appendix_iteration}
Here, we show the results of IterNorm with different iteration numbers on the experiments described in Section 4.2 of the main paper. We also show the results of Batch Normalization (BN) \cite{2015_ICML_Ioffe} and Decorrelated Batch Normalization (DBN) \cite{2018_CVPR_Huang}  for comparison. 

Figure \ref{fig:exp_MLP} shows the results on MNIST dataset.   We explore the effects of $T$ on performance of IterNorm, for a range of $\{1, 3, 5, 7, 9\}$. We observe that the smallest (T = 1) and the largest (T = 9) iteration number both have the worse performance in terms of training efficiency. Further, when T = 9, IterNorm has significantly worse test performance. These observations are consistent to the results on VGG network described in Section 5.1 of the main paper.

Figure \ref{fig:exp_SND} shows the results of SND and conditioning analysis. We observe that IterNorm has better conditioning and increasing SND, with increasing iteration $T$. The results show that the iteration number $T$ can be effectively used to control the extent of whitening,  therefore to obtain a good trade-off between the improved conditioning and introduced stochasticity. 

\vspace{-0.05in}
\section{Details of the VGG Network}
\label{sec:appendix_VGG}

As introduced in Section 5.1 of the main paper, we use the VGG networks \cite{2014_CoRR_Simonyan} tailored for $32 \times 32$ inputs. Table \ref{table:VGG} shows the details of the used VGG network.

{\small
\bibliographystyle{ieee_fullname}
\balance
\bibliography{whitening}
}

\end{document}